\documentclass[runningheads]{llncs}
\usepackage{multirow}
\usepackage{graphicx}
\usepackage{array}

\newcolumntype{P}[1]{>{\centering\arraybackslash}p{#1}}

\usepackage{amsmath}
\usepackage{amssymb}
\usepackage{xcolor}
\usepackage[normalem]{ulem}
\usepackage{minted} 
\usepackage{soul}
\usepackage{makecell}

\begin{document}
\title{A Human-Centered Validation of the Explainability-Performance Coefficient}

\titlerunning{A Human-Centered Validation of the EPC}
%

\author{Christian Oliva\orcidID{0000-0002-8785-6252} \and Luis F. Lago-Fernández\orcidID{0000-0001-8639-8731}}
\authorrunning{Oliva, C. and Lago-Fernández, L.F.}
%

\institute{Grupo de Neurocomputación Biológica, Departamento de Ingeniería Informática, Escuela Politécnica Superior, Universidad Autónoma de Madrid, Spain
\email{christian.oliva@uam.es,}
\email{luis.lago@uam.es}}

\maketitle              

\begin{abstract}

The rapid adoption of deep learning models in high-risk domains has intensified the need for trustworthy Explainable Artificial Intelligence (XAI). However, objectively evaluating explanation fidelity and aligning XAI metrics with human-centered understanding remain critical open challenges. In this work, we propose a model-agnostic metric, the EPC score, which is an extension of the Explainability-Performance Coefficient (EPC), that quantifies explanation quality by explicitly balancing the trade-off between feature selection sparsity and preserved model performance. Through an empirical validation across tabular, text, and image modalities, we show that the EPC score effectively uncovers operational dependencies among network activations, data dimensionality, and explainer performance. Furthermore, we validate the EPC score against independent human-based explanations, proving that higher EPC scores strongly align with human lexical sentiment judgments and spatial visual annotations. 


\end{abstract}

\section{Introduction}
\label{sec:introduction}

The increasing adoption of Deep Learning (DL) models across a wide range of application domains, which have achieved state-of-the-art performance in numerous applications due to their ability to learn highly complex representations, has raised significant concerns regarding their lack of transparency and interpretability \cite{lipton_2018}. As these models are often deployed in high-stakes and regulated environments, understanding their predictions has become a critical requirement. For instance, regulatory frameworks such as the General Data Protection Regulation (GDPR) \cite{GDPR} establish the right to obtain explanations for automated decisions, highlighting the need for interpretable models. This need has driven the development of eXplainable Artificial Intelligence (XAI) methods \cite{ABUSITTA2024124710}, which aim to provide human-interpretable insights into model behavior. As a result, a well-known trade-off has emerged between predictive performance and explainability in DL systems \cite{ExplainableAI_Linardatos_2020}.

In the context of neural networks, many approaches focus on generating feature relevance scores or saliency maps, particularly for image-based tasks using Convolutional Neural Networks (CNNs). These methods attempt to highlight the input regions that most influence the model predictions. However, a major challenge is that different explainability methods often yield distinct results regarding which inputs are considered relevant. Furthermore, several studies have shown that such explanations can be unstable, sensitive to noise, and prone to human interpretation biases, creating the illusion of meaningful reasoning while failing to accurately reflect the true decision-making process of the model \cite{lipton_2018}.

In this scenario, evaluating the quality of these explanations remains an open challenge. Existing evaluation approaches typically rely on proxy measures such as fidelity, stability, or robustness \cite{alvarezmelis2018,ghorbani2018}. However, these metrics often capture only partial aspects of explanation quality and may fail to reflect the practical usefulness of explanations. In particular, many current metrics do not explicitly account for the trade-off between the explainability of selected features and their impact on model performance. To address this, recent literature has shifted towards evaluating explanations through insertion and deletion curves \cite{gomez2022,petsiuk2018}. The trade-off between feature sparsity levels and model performance has been evaluated via the Explainability-Performance Coefficient (EPC) \cite{oliva25}, a metric that quantifies the quality of different explanations for the same model by evaluating the trade-off between accuracy and explainability. 

In this work, we argue that a meaningful explanation should not only highlight relevant features but also preserve the predictive capability of the original model when restricted to those features. Inspired by \cite{oliva25}, we contend that explainability and performance are inherently coupled dimensions that should be evaluated together. Nevertheless, the EPC itself has been studied in a limited manner, leaving its consistency across different data modalities, architectures, and evaluation artifacts largely unexplored.

To address this limitation, in this work, we present a comprehensive empirical exploration of the EPC to assess the quality of explanations produced by XAI algorithms. Unlike model-specific evaluation criteria, as a model- and method-agnostic metric, the EPC operates at the level of explanation methods by leveraging their feature relevance outputs, making it applicable to both local and global explanations. The core of our proposal lies in evaluating the EPC curve, providing a comprehensive assessment of how effectively an explanation identifies a subset of features that preserves the predictive performance of the original model across different sparsity levels. Building upon this analysis, we further introduce the EPC score, a scalar summary of the EPC curve that enables straightforward quantitative comparisons between explanation methods while preserving the information in the complete curve.

The main objective of this work is to explore the capabilities of the EPC score across a diverse suite of benchmarks and network architectures, with the aim of verifying whether it can serve as a robust metric capable of distinguishing faithful explanations from superficial or artifact-driven attribution maps. Beyond evaluating the EPC itself, we show that the resulting EPC score correlates with independent human-centered measures of explanation quality. Furthermore, our experiments consistently identify Integrated Gradients \cite{sundararajan_2017} as the strongest overall performer across the evaluated benchmarks. Nevertheless, our results show that explanation fidelity is intrinsically tied to the specific data modality and the network architecture, and the quality of an explanation depends on the operational conditions of the explainer itself.

To this end, we explore the integration of diverse attribute nullification strategies in high-dimensional image processing, revealing how advanced strategies like Gaussian blurring mitigate out-of-distribution artifacts that typically bias visual evaluations on complex datasets like ImageNet \cite{recht2019imagenet}. We also extend this to the temporal domain in Natural Language Processing (NLP) by analyzing token relevance patterns in text sequences. 

We validate the effectiveness and generality of our framework across multiple data modalities, including tabular data, image classification, and sentiment analysis. Experimental results demonstrate that the EPC provides consistent, discriminative evaluations, exposing how specific XAI methods maintain explanation fidelity where traditional approaches fail. The main contributions of this work are as follows:

\begin{itemize}
    \item \textbf{Proposal of the EPC score:} We extend the Explainability-Performance Coefficient (EPC) \cite{oliva25} into a single, standardized scalar metric: the EPC score. We validate its capacity to measure the explicit trade-off between feature explainability and model performance across diverse data modalities (tabular, vision, and text) and architectures (MLPs, CNNs, LSTMs). This metric enables direct, quantitative, and fair comparisons among any explainer, and identifies that Integrated Gradients (IG) consistently outperform simpler gradient-based and model-agnostic explainers.
    
    \item \textbf{Validation against Human-Centered Explanations:} We validate the proposed EPC score against independent human lexical resources (AFINN lexicon \cite{AFINN_Nielsen2011} in NLP) and spatial annotations (\textit{ImageNet} regions of interest \cite{object_localization_ILSVRC15} in Computer Vision), proving that higher EPC scores strongly align with human-centered explanations.
\end{itemize}

The remainder of this paper is organized as follows: Section~\ref{sec:related-work} establishes the theoretical background and reviews existing evaluation metrics for local explainability. Section~\ref{sec:EPC} formalizes the mathematical framework of the Explainability-Performance Coefficient and introduces the proposed EPC score, detailing feature selection and nullification strategies. Section~\ref{sec:XAI-methods} provides a detailed review of the local explainers evaluated in this work, categorizing them into model-agnostic and model-specific approaches. Section~\ref{sec:setup} describes the experimental setup, detailing the datasets, neural architectures, nullification procedures, and the proposed experiments. Section~\ref{sec:results} presents the experimental results and discussion, structured according to computational feasibility, explainer robustness across activation functions, and human-centered validation in text (IMDB) and image classification (\textit{ImageNet}). Finally, Section~\ref{sec:conclusions} presents our conclusions and outlines directions for future work.

\section{Background and Related Work}
\label{sec:related-work}

This section establishes the theoretical and methodological foundations required to evaluate explainability methods in Deep Learning. In Section \ref{subsec:taxonomy}, we begin by clarifying the taxonomy and foundational definitions of interpretability and explainability, focusing particularly on the distinction between local and global approaches. Next, in Section \ref{subsec:xai_methods}, we review feature attribution methods widely deployed in deep vision and sequential architectures, categorizing them into perturbation-based, gradient-based, and propagation-based techniques. Finally, in Section \ref{subsec:evaluating_xai_methods}, we analyze contemporary evaluation methods, highlighting some limitations of current proxy metrics and establishing the main motivation for our proposed metric.

\subsection{Definitions and Taxonomy of Explainability}
\label{subsec:taxonomy}

The definition of concepts such as interpretability, explainability, understandability, and transparency remains an open issue. Although there is no general consensus and different authors propose subtle differences in their precise definitions \cite{Gilpin2018}, these concepts are often related to the question \textit{``Why does this model make this particular decision?''} \cite{Bromberger1992,Thagard1978}.

In general, two main perspectives are commonly considered to address this question. On the one hand, feature-based explanations focus on identifying the subset of input features that most influence the model's output. On the other hand, model interpretation aims at understanding the internal mechanisms of the model, including neurons, layers, and parameters. Although conceptually different, both approaches are often used interchangeably in the literature and are considered complementary in practice \cite{ExplainableAI_Linardatos_2020,lipton_2018}. In this work, we adopt the convention in which \textit{explainability} refers to feature-based approaches, while \textit{interpretability} refers to internal model understanding. We use the term \textit{transparency} as an umbrella concept encompassing both perspectives.

Beyond the nature of the explanation, explainability methods can also be categorized based on their operational scope into global and local approaches. \textit{Local explainability} focuses on a single instance, uncovering the precise rationale behind a specific prediction by quantifying the contribution of each input feature for that particular sample. On the other hand, \textit{global explainability} aims to provide a general understanding of the model's overall logic, describing how features influence predictions across the entire dataset without focusing on a single instance. In this work, we focus specifically on local explainability methods. 

\subsection{Local Explainability Methods in Deep Learning}
\label{subsec:xai_methods}

Explainability methods in deep learning aim to identify the input attributes that most influence model predictions. These methods are typically implemented through feature importance scores, saliency maps, or attention mechanisms \cite{ExplainableAI_Linardatos_2020}.

A broad distinction can be made between \textit{model-specific} and \textit{model-agnostic} explainability methods. Model-specific approaches are tightly coupled to the internal structure of a given architecture, such as gradient-based methods or propagation-based techniques. In contrast, model-agnostic methods operate independently of the underlying model architecture, treating the model as a black-box and relying solely on input-output behavior. Representative examples of this category include SHAP \cite{Lundberg_SHAP} and LIME \cite{ribeiro_LIME}, which estimate feature contributions by approximating local or global surrogate models.

Among feature attribution approaches, perturbation-based methods evaluate the importance of input regions by modifying or occluding parts of the input and measuring the resulting change in predictions \cite{zeiler_2014_visualizing}. Gradient-based approaches, in contrast, exploit the sensitivity of the output with respect to the input, where gradients directly quantify how small changes in the input affect the prediction \cite{Simonyan2013DeepIC}. Within this category, techniques like Gradient $\times$ Input \cite{Shrikumar_2017} scale the raw gradient by the input values to improve contrast. To resolve the saturation drawbacks of pure gradient methods, Integrated Gradients (IG) \cite{sundararajan_2017} computes the path integral of gradients along a straight line from a baseline reference to the input instance. Alternatively, for convolutional architectures, Grad-CAM \cite{gradcam} leverages the coarse semantic gradients flowing into the final convolutional layer to produce coarse localized saliency maps that focus on high-level visual features.

Another relevant class of methods is based on propagation rules, such as Layer-wise Relevance Propagation (LRP) \cite{bach_lrp}, which redistributes the prediction backward through the network using specific conservation principles. LRP has been shown to produce faithful explanations, although its behavior depends on the choice of propagation rules across layers \cite{Lapuschkin_2017,Montavon2019}. Notably, under certain architectural constraints, the propagation rules of LRP can be reformulated in terms of gradient computations \cite{Ancona2017TowardsBU}. This equivalence highlights that while LRP operates via layer-by-layer redistribution, its local explanations inherently capture the same information as gradient-based attributions.

Despite the large number of proposed methods, most practical applications in computer vision and natural language processing rely primarily on feature attribution techniques, particularly saliency maps, due to their simplicity and interpretability. In addition, model-agnostic approaches such as SHAP and LIME are widely used when model access is limited or when a unified explanation framework is required across heterogeneous architectures.

\subsection{Evaluation of Explainability Methods}
\label{subsec:evaluating_xai_methods}

Evaluating the quality of explanations remains an open challenge in the field of explainable artificial intelligence. Existing approaches typically rely on proxy metrics such as fidelity, robustness, or stability \cite{alvarezmelis2018,ghorbani2018}. Fidelity measures how accurately an explanation reflects the true internal decision-making process of the black-box model. Robustness assesses the explanation's resilience against adversarial manipulations or structural changes in the model, ensuring that similar models yield coherent explanations. Meanwhile, stability quantifies the invariance of the explanation when faced with minor, non-semantic perturbations in the input data, meaning that two almost identical data should produce nearly identical explanations. While useful, these metrics often capture only partial aspects of explanation quality and do not necessarily reflect whether the explanation identifies features that are truly relevant to the predictive behavior of the model.

A significant advancement in this area is the use of causal perturbation-based frameworks, popularized by Petsiuk et al. \cite{petsiuk2018}. In addition to the RISE algorithm, a perturbation-based and model-agnostic explainability method for image processing, they proposed an evaluation of the quality of an explanation based on insertion and deletion curves. These curves quantify the change in model confidence when pixels are added to or removed from the input, following the relevance provided by the explanation. The performance is typically summarized by the Area Under the Insertion or Deletion Curve (IAUC or DAUC) \cite{petsiuk2018}. However, while these metrics provide a causal link between features and predictions, they are often sensitive to the choice of nullification method (blurring vs. constant masking) and do not always account for the global trade-off between model performance and explanation sparsity.

In a similar way, the Explainability-Performance Coefficient (EPC) \cite{oliva25} proposes an alternative approach for quantifying the quality of an explanation. This provides a model- and method-agnostic metric that measures the trade-off between accuracy and explainability. In this work, we argue that a meaningful explanation should not only highlight relevant features but also preserve the predictive capability of the original model when restricted to those features. Thus, we extend the work by Oliva and Lago-Fernández \cite{oliva25} across different data modalities, architectures, and evaluation artifacts, by introducing the EPC score, a scalar summary of the EPC curve that enables quantitative comparisons between explanation methods (hereafter, \textit{explainers}).

\section{The Explainability-Performance Coefficient}
\label{sec:EPC}

This section details the formal mathematical framework of the Explainability-Performance Coefficient (EPC) \cite{oliva25} proposed to evaluate local explanations. We begin by establishing the problem formulation in Section \ref{subsec:problem-formulation}, defining the model behavior, the relevance vectors, and the nature of the perturbation functions across different data modalities. Next, in Section \ref{subsec:feature-selection}, we describe the thresholding and feature selection strategies used to segment explanation rankings into disjoint positive and negative feature subsets. Finally, in Section \ref{subsec:definition}, we describe the EPC and present the EPC score, a single scalar metric for analyzing how the EPC serves as a diagnostic tool for explanations.

\subsection{Problem Formulation}
\label{subsec:problem-formulation}

Let $M : \mathbf{X} \rightarrow \mathbf{Y}$ be a machine learning model that maps an input space $\mathbf{X} \subseteq \mathbb{R}^d$ to a set of predictions $\mathbf{Y}$. For a given input instance $\mathbf{x} \in \mathbf{X}$, the model produces an output $\mathbf{y} = M(\mathbf{x})$, which typically represents a vector of class probabilities or continuous values.

A local explanation algorithm $A$ aims to provide insights into the decision-making process of $M$ by assigning a relevance score to each input feature. Formally, for a specific instance $\mathbf{x}$, $A$ generates a relevance vector $R_\mathbf{x} \in \mathbb{R}^d$:

\begin{equation}
    R_\mathbf{x} = A(M, \mathbf{x}) = (r_1, r_2, ..., r_d),
\end{equation}

\noindent where each $r_i$ quantifies the contribution of the $i$-th feature to the prediction $M(\mathbf{x})$. In this work, we consider a signed relevance framework where:

\begin{itemize}
    \item Positive relevance ($r_i > 0$) indicates that the feature supports the model's current prediction. Removing or nullifying such a feature is expected to decrease the model's confidence in the target class.
    \item Negative relevance ($r_i < 0$) indicates that the feature contradicts the prediction or supports an alternative class. Nullifying these features is expected to increase the model's confidence in the target class.
\end{itemize}

The core objective of the EPC is to quantify the equilibrium between the fraction of nullified features and the resulting impact on the model's performance. To achieve this, we rely on a perturbation function $\Phi(\mathbf{x}, \mathcal{S})$, which nullifies a subset of features ($\mathcal{S}$). 

The nullification is executed using a baseline value or transformation. For low dimensional or normalized datasets, $\Phi$ represents a mean imputation (zero-masking in normalized space), whereas for high-dimensional visual or textual data, $\Phi$ could also denote a local blurring operator or padding, respectively, to mitigate the introduction of high-frequency artifacts that could bias the model's response \cite{petsiuk2018}. By systematically selecting different subsets $S$ based on the relevances $R_\mathbf{x}$, we can evaluate the causal link between the explanation and the model's logic through the lens of predictive consistency.

\subsection{Feature Selection from Explanations}
\label{subsec:feature-selection}

Given a relevance vector $R_\mathbf{x} \in \mathbb{R}^d$, we map the raw scores into binary masks to perform perturbation experiments. To ensure a fair comparison across different explainers, which may produce relevance scores with varying scales and distributions, we employ a percentile-based thresholding strategy. 

Let $F$ be the set of all input features. For a given fraction $k \in [0, 100]$, let $U_k$ denote the relevance threshold such that the bottom $k\%$ of features have relevance values lower than or equal to $U_k$, while the remaining $(100-k)\%$ have relevance values greater than $U_k$. This threshold divides the input features into two disjoint subsets:

\begin{itemize}
    \item Top-$k$ selection ($\mathcal{S}_{top}^k$): Features with high positive relevance, defined as $\{i \in F \mid r_i > U_{k}\}$. These features are hypothesized to be the primary drivers of the model's predictions.
    \item Bottom-$k$ selection ($\mathcal{S}_{bottom}^k$): Features with lower or negative relevance, defined as $\{i \in F \mid r_i \leq U_k\}$. These features represent information that the model effectively ignores or treats as contradictory to the current prediction.
\end{itemize}

\begin{figure}
    \centering
    \includegraphics[width=0.7\linewidth]{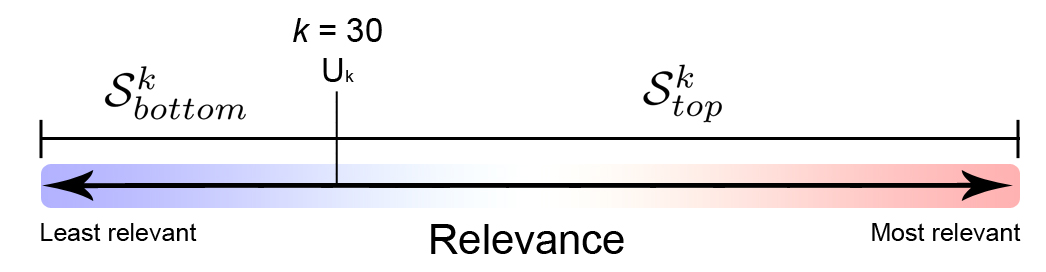}
    \caption{Illustration of the percentile-based feature partitioning strategy. For a given fraction $k$, the feature space is divided into the subset of the $k\%$ least relevant features ($\mathcal{S}_{bottom}^k$) and the complementary subset containing the remaining $(100-k)\%$ most relevant features ($\mathcal{S}_{top}^k$).}
    \label{fig:illustrative_subsets}
\end{figure}

For any fixed fraction $k$ and the corresponding threshold $U_k$, these subsets are mutually exclusive and collectively exhaustive, satisfying $\mathcal{S}_{top}^k \cap \mathcal{S}_{bottom}^k = \emptyset$ and $\mathcal{S}_{top}^k \cup \mathcal{S}_{bottom}^k = F$. This partitioning is illustrated in Figure \ref{fig:illustrative_subsets}, which visually shows how varying the fraction $k$ selects a subset containing the $k\%$ least relevant features ($\mathcal{S}_{bottom}^k$) and a complementary subset containing the remaining $(100-k)\%$ most relevant features ($\mathcal{S}_{top}^k$).

While literature often evaluates XAI frameworks using both insertion (building an image from a baseline) and deletion (removing features from the original image) protocols \cite{petsiuk2018}, these two paradigms represent complementary perspectives of the same underlying causal attribution. Therefore, to ensure clarity and avoid redundancy, our evaluation framework focuses strictly on the deletion regime through two distinct scenarios:

\begin{itemize}
    \item Deletion top-$k$: Measuring performance degradation by nullifying the most positive features, $\Phi(\mathbf{x}, \mathcal{S}_{top}^k)$. A faithful explanation should trigger a rapid drop in model confidence, as key evidence is removed.
    \item Deletion bottom-$k$: Measuring performance stability by nullifying the most negative features, $\Phi(\mathbf{x}, S_{bottom}^k)$. A robust explanation should keep model confidence intact, as only non-essential or contradictory features are discarded.
\end{itemize}

\subsection{EPC Definition and the EPC Score}
\label{subsec:definition}

The Explainability-Performance Coefficient (EPC), which was previously introduced in \cite{oliva25}, is defined as a metric that quantifies the discriminative power of an explanation $R_\mathbf{x}$ at a specific sparsity level $k$. It measures the ``performance gap" between selecting features that the explainer deems relevant versus those it deems irrelevant. The EPC is defined as follows:

\begin{equation}
    EPC(R_\mathbf{x}, k, M) = \frac{k}{100} \times \frac{P(M, \Phi(\mathbf{X}, \mathcal{S}_{bottom}^{k})) - P(M, \Phi(\mathbf{X}, \mathcal{S}_{top}^{k}))}{P(M,\mathbf{X})}, \label{eq:EPC}
\end{equation}

\noindent where:

\begin{itemize}
    \item $P(M, \Phi(\mathbf{X}, \mathcal{S}_{bottom}^k))$ represents the model's performance when we retain the $(100-k)\%$ most relevant features (by nullifying the $k\%$ least relevant ones, $\mathcal{S}_{bottom}^{k}$).
    \item $P(M,\Phi(\mathbf{X}, \mathcal{S}_{top}^{k}))$ represents the model's performance when we retain the $k\%$ least relevant features (by nullifying the $(100-k)\%$ most relevant ones, $\mathcal{S}_{top}^{k}$).
    \item $P(M,\mathbf{X})$ is the original performance score, acting as a normalizer.
\end{itemize}

The EPC value provides a snapshot of the explanation's reliability at a specific level of information compression $k$. A positive EPC indicates that the features identified as relevant by the explainer are indeed more influential for the model's prediction than the set of features identified as non relevant. On the other hand, a zero or negative EPC indicates a failure in the explanation. A zero value implies that the relevant features are indistinguishable from the irrelevant ones in terms of the model performance, while a negative value suggests that the explanation is misidentifying noise or contradictory features as primary drivers of the prediction. By evaluating the EPC at different $k$ values, we can compare how the explainers prioritize the most critical information for the model.

Under its formulation (Eq. \ref{eq:EPC}), the expected maximum upper bound for a given $k$ occurs when the explainer achieves perfect discrimination. This means that the model retains maximum performance when preserving the top-ranked features, that is $P(M, \Phi(\mathbf{X}, \mathcal{S}_{bottom}^k)) = P(M, \mathbf{X})$, and it collapses completely when they are omitted, $P(M, \Phi(\mathbf{X}, \mathcal{S}_{top}^{k})) = 0$. Consequently, the expected optimal EPC describes a linear reference ceiling given by $f(k) = k / 100$ since the second term in Eq. \ref{eq:EPC} is equal to $1$.

Considering this reference ceiling $f(k)$, we seek a metric that summarizes the EPC into a single scalar. By calculating the area between $f(k)$ and the EPC curve, $EPC(k)$, we can condense the global quality of the explanation. We normalize this area with respect to the area under $f(k)$, and subtract this value from one to obtain the EPC score $E$:

\begin{equation}
    E = 1 - \frac{\Delta k \sum_k (f(k) - EPC(k))}{50} \label{eq:epc_score}
\end{equation}

This formulation yields a bounded score where higher values indicate stronger alignment with $f(k)$. In particular, values close to $1$ correspond to near-perfect alignment across sparsity levels, while lower values reflect increasing deviation from the optimal trajectory. Notably, negative values may arise in this score when the EPC curve reaches negative values.

In addition, we must consider that nullifying non-relevant attributes may lead to an increase in model performance. This behavior occurs because removing non-relevant attributes, which support the prediction of competing classes, can help the model make more accurate predictions. Such improvements are more likely when the baseline model has relatively low performance. In this scenario, the EPC curve may exceed $f(k)$, and thus the score $E$ can be greater than $1$.

Since the EPC relies exclusively on input perturbations and the observation of the resulting outputs, it is strictly model-agnostic. It can be applied to evaluate explanations from any machine learning model. Unlike traditional fidelity metrics that only look at one side of the explanation, the EPC introduces a contrastive mechanism. It evaluates the explanations and shows whether the top-ranked features are more useful within the model's logic than the bottom-ranked ones. The use of top-ranked features measures the necessity of these features for maintaining the prediction, while the use of bottom-ranked features prevents the metric from overestimating the quality of explanations in models that are not robust to information loss. 

In addition, the EPC is sensitive to the choice of the perturbation function $\Phi$. While it does not dictate a specific nullification method, the resulting score reflects how the model's performance is affected by different strategies (e.g., mean imputation vs. Gaussian blur). This property is particularly relevant for high-dimensional datasets, where the choice of the baseline can significantly influence the model's latent representations.

\section{Review of Local Explainers}
\label{sec:XAI-methods}

To evaluate the EPC, we benchmark a diverse selection of state-of-the-art local explainers. These techniques are fundamentally divided into model-agnostic approaches (see Section \ref{subsec:model-agnostic-methods}), which treat the architecture as a black-box, and model-specific methods (described in Section \ref{subsec:model-specific-methods}), which leverage internal gradients and network architecture.

\subsection{Model-agnostic Methods}
\label{subsec:model-agnostic-methods}

Model-agnostic methods operate independently of the underlying neural network architecture, relying strictly on systematic perturbations of the input space and the observation of the corresponding marginal changes in the model's output.

\subsubsection{SHapley Additive exPlanations (SHAP).}

SHAP \cite{Lundberg_SHAP} is grounded in cooperative game theory, framing the attribution problem as the allocation of prediction shifts among input features. It computes the unique additive feature importance values, known as Shapley values $\phi_i$, which satisfy desirable axioms such as efficiency, symmetry, and dummy allocation. Due to its combinatorial complexity, SHAP requires sampling approximations (e.g., KernelSHAP), making it highly accurate but computationally expensive in high-dimensional spaces.

\subsubsection{Local Interpretable Model-agnostic Explanations (LIME).}

LIME \cite{ribeiro_LIME} explains individual predictions by training an interpretable model $G$ (such as a linear regressor) to predict the output of the explained model $M$ in the local neighborhood of a target instance $\mathbf{x}$. The coefficients $\beta$ of this linear model represent the marginal effects, serving as the local feature relevances that determine the direction and rate of change in the model's predictions.

Since these coefficients only capture local variations, similarly to gradients, we follow the idea of Shrikumar et al. \cite{Shrikumar_2017} (see Section \ref{subsec:model-specific-methods}) and introduce a variant that we term LIME $\times$ Input, where the relevance of attribute $x_i$ is defined as $R_i = \beta_i \cdot x_i$.

\subsection{Model-specific Methods}
\label{subsec:model-specific-methods}

Model-specific methods leverage the internal structural properties of the neural network, such as architecture, layer weights, activations, and backpropagated gradients, to compute feature relevance maps.

\subsubsection{Gradient $\times$ Input.}
Based on foundational saliency maps \cite{Simonyan2013DeepIC}, Gradient $\times$ Input \cite{Shrikumar_2017} scales the raw gradient of the output with respect to the input by the magnitude of the input values themselves. This operation aims to improve visual contrast and account for feature scale. The relevance $R_i$ for each feature is computed as:

\begin{equation}
    R_i^{Grad\times Input} = x_i \cdot \frac{\partial M(\mathbf{x})}{\partial x_i} \label{eq:gradientxinput}
\end{equation}

\subsubsection{Layer-wise Relevance Propagation (LRP).}

LRP \cite{bach_lrp} operates via a backward propagation pass governed by conservative redistribution rules. The simplest formulation, LRP-0, redistributes the total relevance $R_j$ from an upper-layer neuron $j$ to a lower-layer neuron $i$ according to the ratio of their forward activations and connection weights:

\begin{equation}
    R_i^{LRP} = \sum_j \frac{a_i w_{ij}}{\sum_k a_k w_{kj}} R_j,
\end{equation}

\noindent where $a_i$ is the activation of neuron $i$, and $w_{ij}$ is the weight connecting neuron $i$ to neuron $j$. A well-known result in the XAI literature demonstrates that LRP-0 can be efficiently implemented via a modified backpropagation pass, where the standard derivative of the activation function $f(x)$ is replaced by the ratio $f(x)/x$ \cite{Ancona2017TowardsBU}.

Despite its efficiency in ReLU-based networks, the gradient-equivalent implementation of LRP-0 \cite{Ancona2017TowardsBU} faces critical theoretical and practical challenges in more complex architectures that incorporate non-linear activations like the Sigmoid ($\sigma$) or hyperbolic tangent ($\tanh$). These are still used in Recurrent Neural Networks (RNNs) and gated architectures (such as LSTMs or GRUs) for processing sequential data, while they are less frequent in standard feedforward or convolutional layers. In layers utilizing the Sigmoid function, the ratio $\sigma(x)/x$ becomes numerically unstable as $x$ approaches $0$. This causes the pseudo-gradient term to approach infinity, exploding the relevance scores and rendering LRP-0 mathematically unusable.

\subsubsection{Integrated Gradients (IG).}

To satisfy the axiom of implementation invariance and overcome saturation, Integrated Gradients \cite{sundararajan_2017} aggregates the gradients along a straight path from a user-defined reference baseline $x'$ to the input instance $x$. The relevance is defined as:

\begin{equation}
    R_i^{IG} = (x_i - x'_i) \cdot \int_{0}^{1} \frac{\partial M(\mathbf{x}' + \alpha(\mathbf{x} - \mathbf{x}'))}{\partial x_i} d\alpha, \label{eq:IG}
\end{equation}

By averaging the gradients across varying intensities ($\alpha$), IG captures the importance of saturated features that become active along the path. In our experiments, we approximate this integral using a Riemann summation with $m=20$ steps.

\subsubsection{Linear Integrated Gradients (LIG).}

As a computationally streamlined alternative to the full path integral, we also evaluate a single-step path linear approximation, which we denote as Linear Integrated Gradients (LIG). This approach serves as an intermediate step between Gradient $\times$ Input and IG. By setting $\alpha=1$ directly inside the integral (Eq. \ref{eq:IG}) and utilizing a single Riemann step ($m=1$), the method takes a single gradient evaluation at the target instance $\mathbf{x}$ relative to the direct difference from the baseline $\mathbf{x}'$:

\begin{equation}
    R_i^{LIG} = (x_i - x'_i) \cdot \frac{\partial M(\mathbf{x})}{\partial x_i}
\end{equation}

This single-step approximation aligns with the mechanics conceptualized by Shrikumar et al. \cite{Shrikumar_2017} when evaluating attributions relative to a reference baseline. It is worth noting that when utilizing a standard zero-baseline ($\mathbf{x}'=\mathbf{0}$), LIG is the classic Gradient $\times$ Input (Eq. \ref{eq:gradientxinput}).

\subsubsection{Gradient-weighted Class Activation Mapping (Grad-CAM).}

Specifically tailored for convolutional neural networks (CNNs) in visual tasks, Grad-CAM \cite{gradcam} calculates coarse localization maps. It computes the gradient of the score for class $c$ ($y^c$) with respect to the feature map activations $A^k$ of the final convolutional layer. These gradients are globally pooled to capture the importance weight $\alpha_k^c$ of each feature map:

\begin{equation}
    \alpha_k^c = \frac{1}{Z} \sum_i \sum_j \frac{\partial y^c}{\partial A_{ij}^k},
\end{equation}

\noindent where $Z$ is the spatial area of the feature map. A weighted combination of forward activation maps is followed by a ReLU operation to retain only features that positively correlate with the target class:

\begin{equation}
    L_{Grad-CAM}^c = \text{ReLU}\left(\sum_k \alpha_k^c A^k\right).
\end{equation}

Since $L_{Grad-CAM}^c$ is a coarse, low-resolution heatmap matching the spatial dimensions of the final convolutional feature maps, it cannot be directly applied to instance-level pixel masking. To align the explanation with the input space for EPC evaluation, the final relevance vector $R_{ij}^{Grad-CAM}$ is obtained by projecting $L_{Grad-CAM}^c$ back to the original image dimensions $(H \times W)$ using a bilinear interpolation transformation.

\section{Experimental Setup}
\label{sec:setup}

This section outlines the comprehensive empirical framework designed to evaluate the performance and structural properties of the EPC. We first introduce in Section \ref{subsec:datasets} the datasets and data modalities spanning tabular records, computer vision, and sequential natural language processing. Next, in Section \ref{subsec:models}, we detail the target architectures and specific training configurations. We then formalize in Section \ref{subsec:nullification} the concrete implementation details and the specific attribute nullification strategies deployed to alter feature spaces. Finally, in Section \ref{subsec:roadmap}, we map the experimental setup, establishing the three core evaluation objectives that guide the empirical analysis of this work.

\subsection{Datasets and Data Modalities}
\label{subsec:datasets}

To demonstrate the generality of our framework, we evaluate it across three distinct data modalities. For each modality, explanations are computed on a fixed subset of the corresponding training dataset, allowing for a consistent and computationally tractable evaluation across experiments.

\begin{itemize}
    \item Tabular data: we use a bank loan default dataset from Kaggle \cite{kaggle_dataset}, consisting of $45.000$ records and 14 features. Since categorical features are converted to one-hot encoding, the resulting data contains $21$ attributes in our experiments. All variables are standardized prior to training, providing a clean, controlled environment to assess the EPC in a low-dimensional feature space where ground-truth relevance is more intuitive. Explanations are evaluated on a subset of $1.000$ training samples.
    \item Image Data: We use the classic \textit{MNIST} \cite{mnist} dataset for digit recognition ($28 \times 28$ grayscale pixels), where explanations are evaluated on a subset of $1.000$ training images, and \textit{ImageNet} \cite{DenDon09Imagenet} (via the \textit{Imagenette} \cite{imagenette} 10-class subset) for complex visual tasks, where all images are resized to a standard resolution of $224 \times 224 \times 3$ pixels. In the context of \textit{ImageNet}, we define $R_\mathbf{x}$ not merely as a spatial pixel $(h, w)$, but as a channel-specific feature $(c,h,w)$. This allows the explanation to reflect the model's sensitivity to color or spectral information, treating each channel as a distinct dimension for relevance assignment. For the computationally expensive XAI evaluation, we sample a fixed subset of $100$ randomly selected images from the dataset. To explore the relationship between the EPC and human visual understandability, we additionally use the \textit{ImageNet} Object Localization annotations \cite{object_localization_ILSVRC15}, available through the \textit{ImageNet} Object Localization Challenge \cite{imagenet-object-localization-challenge}, which provides manually annotated Regions of Interest (ROIs) for the evaluated images. 
    \item Text Data: For Natural Language Processing, we use the \textit{IMDB} sentiment analysis dataset \cite{imdb_dataset}. Sequences are padded to a fixed length of $200$ tokens, considering the $10,000$ most frequent words in the vocabulary. Explanations are evaluated on a subset of $1.000$ training samples. To assess semantic agreement with human lexical knowledge, we complement this dataset with the AFINN sentiment lexicon \cite{AFINN_Nielsen2011}, which assigns manually curated sentiment polarity scores to English words.
\end{itemize}

\subsection{Target Models, Training Architectures, and Hyperparameters}
\label{subsec:models}

To rigorously use the EPC to evaluate different explainability methods under diverse regimes, we deploy a wide spectrum of deep learning architectures. We emphasize that the primary objective of these setups is not to achieve state-of-the-art predictive performance, but rather to analyze how the quality and consistency of XAI explanations adapt to varying model capacities, non-linear activation functions, and architectural complexities. 

\subsubsection{Multilayer Perceptrons (MLPs).}

For both the Kaggle tabular and the MNIST datasets, we implement a streamlined Multilayer Perceptron (MLP) architecture. For MNIST, an MLP is intentionally chosen over a standard Convolutional Neural Network (CNN) for simplicity, allowing us to evaluate pixel-level attribution without the structural inductive biases of spatial convolutions. Both networks consist of a single hidden dense layer with $20$ neurons. 

To systematically test the resilience of our metric within the explanations against gradient saturation and numerical instabilities, we experiment with four distinct hidden activation functions: Rectified Linear Unit (ReLU), Sigmoid ($\sigma$), Hyperbolic Tangent ($\tanh$), and Sigmoid Linear Unit (SiLU). Training is executed over $10$ epochs using the Adam optimizer with a learning rate of $0.001$ and a batch size of $32$. As discussed in Section~\ref{sec:results}, these lightweight environments are also strategically utilized to validate execution bottlenecks, serving to empirically discard SHAP and LIME due to their prohibitive computational costs.

\subsubsection{Convolutional Neural Networks (CNNs).}

For the high-dimensional visual task on ImageNet (via the \textit{Imagenette} subset), our framework evaluates two contrasting architectural scenarios:
\begin{enumerate}
    \item Custom CNN (Weak Model): Built from scratch to test the EPC's sensitivity to sub-optimal classifiers. The architecture is defined sequentially as: $\text{Conv2D}(10, 3\times3) \to \text{MaxPool} \to \text{Conv2D}(10, 3\times3) \to \text{MaxPool} \to \text{Flatten} \to \text{Dense}(20) \to \text{Softmax}(10)$. It is trained for $10$ epochs using Adam (learning rate = $0.001$), a batch size of $128$, and swept across the same four activation functions (ReLU, SiLU, $\tanh$, and Sigmoid). 
    \item MobileNet (High-Performance Model): We leverage a pre-trained off-the-shelf \textit{MobileNet} \cite{mobilenet_2017} architecture to test the scalability of the EPC under highly optimized, complex feature extraction landscapes.
\end{enumerate}

\subsubsection{Recurrent Neural Networks (RNNs).}

For sequential natural language processing on the IMDB sentiment dataset, models are trained for $10$ epochs using Adam (learning rate = $0.001$), and a batch size of $128$. We benchmark a standard recurrent architecture to validate our framework, composed of an $\text{Embedding}(16) \to \text{LSTM}(10) \to \text{Softmax}(2)$ pipeline. 

\subsection{Implementation Details and Nullification Strategies}
\label{subsec:nullification}

The evaluation is conducted using Tensorflow \cite{tensorflow2015} and Keras \cite{chollet2015keras}, with OpenCV \cite{opencv_library} for image transformations. The EPC is calculated by iterating over the sparsity threshold defined by $k$ from $0$ to $100$ with a step size of $\Delta k$. For the Tabular dataset, which only contains 21 attributes, $\Delta k = 100 / 21 \approx 4.76$ (the addition of each attribute corresponds to one evaluation step). For all other datasets, $\Delta k = 1$. 

Regarding the nullification operator $\Phi$, for tabular and MNIST data, we use mean imputation (zero-masking); for \textit{ImageNet}, we compare mean imputation (zero-masking) and Gaussian Blur (used to prevent the model from reacting to high-frequency artifacts induced by sharp masks \cite{dodge_2016}). This Gaussian blur is parameterized with a local kernel of size $15 \times 15$ pixels, where the standard deviation $\sigma$ is automatically derived from the kernel dimensions to match the spatial scale. Following the blurring transformation, pixel values are strictly clipped back to the original $[-1, 1]$ normalization range to guaranty mathematical consistency and avoid input saturation during inference. Lastly, for IMDB, nullification is performed by replacing word tokens with the padding token (zero index).

Finally, we record execution times for all explainers. Due to the high computational cost of model-agnostic methods like LIME and SHAP, these are primarily evaluated on the simplest dataset (Kaggle) and then discarded for larger and more complex datasets.

\subsection{Evaluation Roadmap}
\label{subsec:roadmap}

The pipeline of our experimental evaluation follows three steps: (i) target models are trained from scratch or loaded from pre-trained repositories under the architectural configurations detailed in Section~\ref{subsec:models}; (ii) local explanations are generated for the selected validation instances using the explainers formulated in Section~\ref{sec:XAI-methods}; (iii) the EPC is dynamically computed by sweeping the sparsity threshold $k$ to test the predictive boundaries of the models. 

Through this, our empirical analysis is explicitly structured around three core evaluation objectives, which directly correspond to the results presented in Section~\ref{sec:results}. First, we assess the capacity of the EPC to differentiate between high-quality and poor explanations in low-dimensional spaces while analyzing the execution time overhead to justify the exclusion of sampling-heavy methods in more complex tasks.

Second, we analyze the robustness of the metric across several neural architectures with different activation functions ($\text{ReLU}$ vs. $\text{SiLU}$ vs. $\text{tanh}$ vs. $\text{Sigmoid}$), allowing us to evaluate whether the ranking induced by the EPC remains stable under distinct scenarios.

Finally, we evaluate whether the quality ranking induced by the EPC is also supported by independent human-centered evidence. For text classification, we compare the relevance assigned to words with human-curated sentiment annotations from the AFINN lexicon \cite{AFINN_Nielsen2011}, studying the consistency with lexical agreement. For image classification, we employ \textit{ImageNet} object localization annotations \cite{object_localization_ILSVRC15} to measure how well the explanations concentrate their relevance inside human-annotated Regions of Interest (ROIs). These experiments show that explainers achieving higher EPC also produce explanations that are more aligned with human semantic understanding.

Before transitioning to the empirical results, we remark that since explainability lacks a universal, absolute ground-truth benchmark, it is impossible to declare an isolated explainer as superior. Consequently, the EPC does not attempt to arbitrarily favor one algorithm over another. Instead, the coefficient operates as an objective framework of explanation quality, directly quantifying the mathematical equilibrium achieved when an explainer successfully minimizes the fraction of relevant features while preserving the operational predictive capability of the underlying model. Nevertheless, consistent superiority of an explainer according to this metric, particularly when it aligns with human-centered evaluations, may provide meaningful evidence of its practical quality and reliability.

\section{Results and Discussion}
\label{sec:results}

This section presents the empirical validation of the Explainability-Performance Coefficient (EPC) across diverse data modalities and architectural configurations. To provide a rigorous and structured analysis, the results are organized directly following the core evaluation objectives established in Section \ref{subsec:roadmap}. We first analyze in Section \ref{subsec:rq1} the validation and computational feasibility of the EPC metric, justifying the pre-filtering of computationally prohibitive baselines. Next, in Section \ref{subsec:rq2}, we test the structural robustness of contemporary explainers under different activation functions. 

The last part focuses on validating the EPC from a human-centered perspective. Using the IMDB benchmark, in Section \ref{subsec:results-text}, we analyze whether higher EPC scores correlate with human-curated lexical sentiment knowledge. Finally, we extend this analysis to computer vision in Section \ref{subsec:results-vision} by comparing explanation maps against manually annotated \textit{ImageNet} Regions of Interest (ROIs). 

\subsection{Validation and Computational Feasibility of the Explainers}
\label{subsec:rq1}

We begin our empirical analysis by validating the discriminative capability of the EPC and assessing the computational scalability of the explainers described in Section \ref{sec:XAI-methods}. Aligning with the framework established in Section~\ref{subsec:roadmap}, we evaluate the EPC not as an arbitrary tool to crown a specific explainer, but rather as an objective metric of informational efficiency. It quantifies an explanation's capacity to minimize the required feature density while effectively preserving the underlying predictive performance of the model.

\subsubsection{Computational Overhead.}

To establish a clear feasibility baseline, we show the execution overhead across all seven explainers introduced in Section \ref{sec:XAI-methods} using the low-dimensional Kaggle tabular dataset. We show in Table \ref{tab:tabular-results} the average execution time over 10 different executions. The values shown measure the the total time (in seconds) required to compute the relevances for each input attribute for every sample in the validation dataset.

\begin{table}[!bht]
    \centering
    \caption{Comparison of execution times (seconds) for several explainers on the tabular dataset. Note that Grad-CAM is not present in this table because it is a CNN-specific explainer.}
    \begin{tabular}{crc}
        & \textbf{Method} & \textbf{Execution Time (seconds)}  \\
        \cline{2-3}
        \cline{2-3}
        \multirow{2}{*}{Model-Agnostic} & SHAP & 1717 $\pm$ 31  \\ 
        & LIME $\times$ Input & 106 $\pm$ 4 \\ 
        \cline{2-3}
        \multirow{4}{*}{Model-Specific} & Gradient $\times$ Input & 0.01 $\pm$ 0.0 \\
        & LRP & 0.01 $\pm$ 0.0  \\ 
        & Linear IG & 0.02 $\pm$ 0.0  \\
        & IG & 0.11 $\pm$ 0.02 \\
        \cline{2-3}
    \end{tabular}
    \label{tab:tabular-results}
\end{table}

As observed, model-agnostic methods like LIME and SHAP suffer from prohibitive computational costs. SHAP requires over 28 minutes to evaluate the dataset, while LIME averages nearly 2 minutes. In contrast, neural network specific methods (Gradient $\times$ Input, LRP, LIG, and IG) complete the evaluation in less than one second. This massive discrepancy motivates the exclusion of SHAP in the MNIST experiments and both SHAP and LIME in the high-dimensional ImageNet benchmarks.

\subsubsection{Understanding the EPC.}

\begin{figure}
    \centering
    \includegraphics[width=\linewidth]{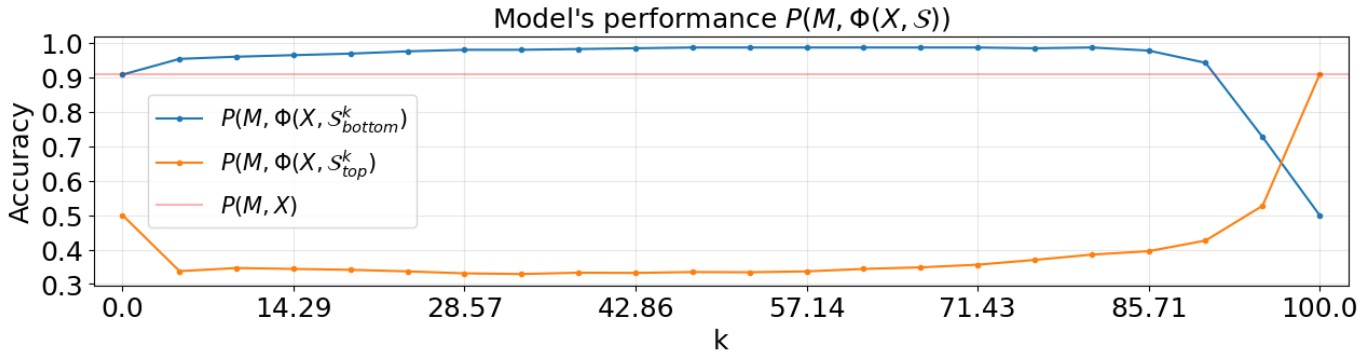}
    \caption{Model performance trajectories when removing non-relevant attributes ($P(M, \Phi(X, \mathcal{S}_{bottom}^k))$, blue curve) and removing relevant attributes ($P(M, \Phi(X, \mathcal{S}_{top}^{k}))$, orange curve) versus the $k\%$ of nullified attributes. Note that nullifying non-relevant attributes leads to an increase in performance (blue curve above the red horizontal line representing the baseline model performance).}
    \label{fig:mmas_vs_mmenos}
\end{figure}

To understand the results, we need to visualize the mechanics behind the EPC. This metric relies on a contrastive evaluation that measures the model performance under two opposing regimes, as described in Equation \ref{eq:EPC}: a progressive selection of highly relevant features, $P(M, \Phi(\mathbf{X}, \mathcal{S}_{bottom}^k))$, and the omission of those same features, $P(M, \Phi(\mathbf{X}, \mathcal{S}_{top}^{k}))$. Figure~\ref{fig:mmas_vs_mmenos} illustrates the typical trajectories of these two metrics when $k$ varies between 0 and 100, computed on the Tabular dataset using Gradient $\times$ Input on an MLP trained with a ReLU activation function. 

We expect that an ideal explanation causes $P(M, \Phi(\mathbf{X}, \mathcal{S}_{bottom}^k))$ (blue curve) to maintain or even increase the baseline performance until the final features, where it is no longer possible to preserve it, as shown in the figure. This indicates that the most critical features have been correctly identified, and also that nullifying counter-relevant features improves the model's performance. Concurrently, $P(M, \Phi(\mathbf{X}, \mathcal{S}_{top}^{k}))$ (orange curve) should maintain low performance until the final values, proving that removing those highly relevant tokens destroys the model's predictive capability. 

Two limit cases for $k=0$ and $k=100$ are worth considering. At $k=0$, no features have been removed from the blue curve, so its performance corresponds to the model's baseline accuracy (red line). In contrast, the orange curve is evaluated after replacing all features with the nullification value (zero in this scenario). Since the Tabular dataset is a balanced classification task, its accuracy is close to $0.5$ because the model predicts always the same class. The opposite behavior occurs at $k=100$: the blue curve reaches $0.5$ accuracy because all features have been replaced by zero, whereas the orange curve recovers the model's baseline accuracy.

\begin{figure}
    \centering
    \includegraphics[width=\linewidth]{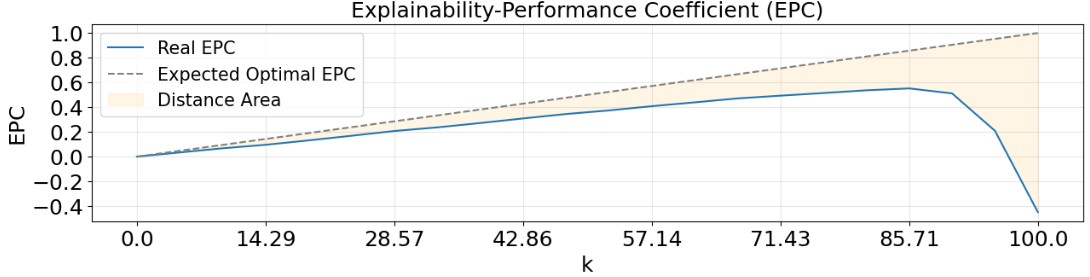}
    \caption{EPC curve (solid blue) plotted against the expected \textit{Optimal EPC} (shaded gray). The shaded region between the two curves defines the distance area between them. A smaller area signifies that the explainer operates closer to the expected limit.}
    \label{fig:epc_vs_optimal}
\end{figure}

The interaction between these two regimes forms the EPC (Eq. \ref{eq:EPC}), which is shown in Figure \ref{fig:epc_vs_optimal} for the Tabular dataset using the same explainer (Gradient $\times$ Input) and model (ReLU activation function) as the ones used in Figure \ref{fig:mmas_vs_mmenos}. As described in Section \ref{subsec:definition}, the expected \textit{Optimal EPC} describes a linear reference ceiling given by $f(k) = k / 100$. Figure~\ref{fig:epc_vs_optimal} compares this reference with the resulting EPC.

\begin{figure}[!bht]
    \centering
    \includegraphics[width=0.32\linewidth]{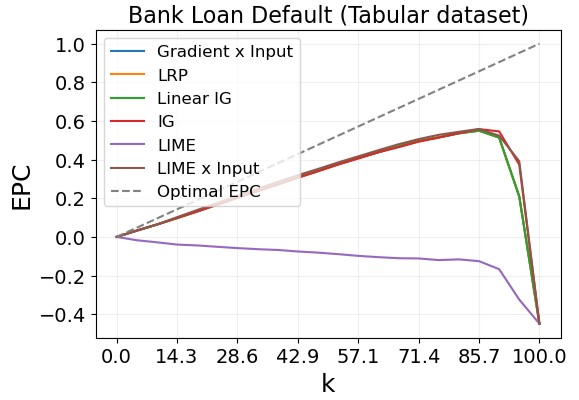}
    \includegraphics[width=0.32\linewidth]{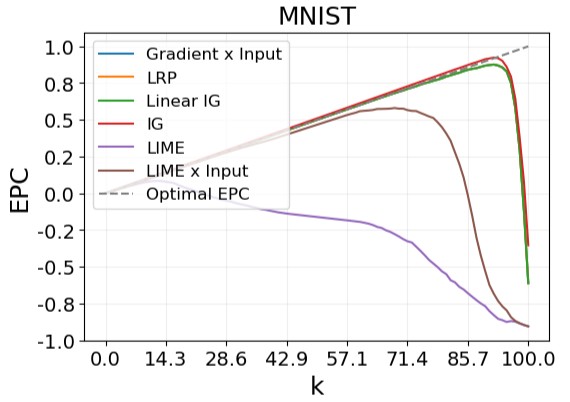}
    \includegraphics[width=0.32\linewidth]{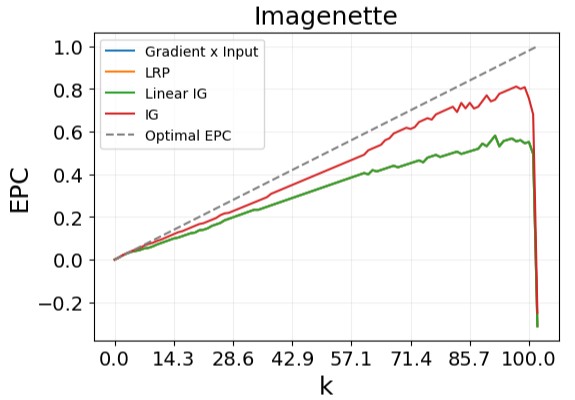}
    \caption{EPC curves for several explainers under ReLU activation function across different data modalities: Tabular dataset (left), MNIST (middle), and custom CNN-Imagenette (right).}
    \label{fig:curves-relu}
\end{figure}

By calculating the area between $f(k)$ and the EPC curve, we can calculate the EPC score (see Eq. \ref{eq:epc_score}). For the example in Figure \ref{fig:epc_vs_optimal}, the EPC score is $E \approx 0.54$. Once this normalized score is established, we analyze various explainers under a standard ReLU activation function. Figure~\ref{fig:curves-relu} presents the EPC trajectories alongside their corresponding expected optimal baselines across three distinct data modalities, while Table~\ref{tab:areas-relu} shows the associated EPC scores.

\begin{table}[!bht]
    \centering
    \caption{EPC scores (higher is better) for each explainer under ReLU activation function. Missing values ($-$) denote configurations discarded due to prohibitive computational costs.}
    \begin{tabular}{rr|c|c|c}
         & \textbf{Explainer} & \textbf{Bank Loan Default} & \textbf{MNIST} & \textbf{Imagenette} \\
         \cline{2-5}
         \cline{2-5}
         \multirow{4}{*}{Model-specific} & Gradient $\times$ Input & 0.54 & 0.91 & 0.65 \\
         & LRP & 0.54 & 0.91 & 0.65 \\
         & LIG & 0.54 & 0.91 & 0.65 \\
         & IG & 0.55 & \textbf{0.95} & \textbf{0.86} \\
         \cline{2-5}
         \multirow{3}{*}{Model-agnostic} & LIME & -0.21 & -0.50 & $-$ \\
         & LIME $\times$ Input & \textbf{0.57} & 0.39 & $-$ \\
         & SHAP & 0.55 & $-$ & $-$ \\
         \cline{2-5}
    \end{tabular}
    \label{tab:areas-relu}
\end{table}

An analysis of the Tabular (Bank Loan Default) dataset reveals that most methods exhibit similar performance, with the EPC scores hovering around $0.55$. Note that Gradient $\times$ Input, LRP, and Linear Integrated Gradients (LIG) yield identical scores ($E = 0.54$) since they are equivalent when the activation is ReLU and zero-baseline is used. In addition, standard LIME underperforms with an EPC score of $-0.21$. Compared with LIME $\times$ Input ($E = 0.57$), it serves as a validation of the necessity of input-sign scaling for nullification-based metrics.

As the data complexity scales to MNIST, the performance gap between model-agnostic and model-specific methods is more evident. Standard LIME breaks down completely, accumulating an EPC score of $-0.50$, while LIME $\times$ Input improves this behavior considerably to $0.39$, though it remains non-competitive against model-specific alternatives. In this case, SHAP is omitted due to execution complexity. Among the model-specific techniques, the mathematical equivalence between Gradient $\times$ Input, LRP, and LIG persists at $0.91$. Nevertheless, multi-step ($m=20$) Integrated Gradients (IG) outperforms all competitors, achieving the highest EPC score of $0.95$.

Finally, for the high-dimensional Imagenette benchmark, all model-agnostic variants are omitted due to the previously discussed execution constraints. The empirical results mirror the hierarchy seen in MNIST: Gradient $\times$ Input, LRP, and LIG remain bound at an identical EPC Score of $0.65$, whereas full multi-step IG showcases its robust architectural advantage ($E = 0.86$).

In summary, these experiments deliver two main conclusions: first, model-agnostic methods scale poorly in both computational efficiency and explanation quality according to the EPC score as data dimensions increase. Second, under standard ReLU structures, simpler gradient approaches (LRP, LIG, and Gradient $\times$ Input) collapse into a single behavioral baseline, while IG emerges as the most effective explainer in this scenario. 

\subsection{Robustness of the Explainers}
\label{subsec:rq2}

In this section, we analyze the structural robustness of local explainability methods when faced with different network architectures and activation functions. As detailed in Section \ref{subsec:models}, for the Tabular and MNIST datasets, we use an MLP with a single hidden dense layer with 20 neurons for simplicity, while for the Imagenette dataset, we use a Custom CNN built from scratch (see details in Section \ref{subsec:models}) to test the EPC sensitivity to sub-optimal classifiers. We alternate now between three additional activation functions: SiLU, tanh, and Sigmoid, to complement the previous results obtained for the ReLU. By shifting these functions, we test how effectively each method maintains its explanation fidelity. 

Since changing the activation function results in different models with different decision boundaries and classification performance, the resulting EPC scores should be interpreted in terms of the relative ranking of explainers within each configuration rather than their absolute values. Thus, EPC scores are only comparable across the explainers for the same model. Table~\ref{tab:area_vs_problem} synthesizes these results by cross-referencing the EPC score across all tested activation functions, data modalities, and explainers.

\begin{table}[!bht]
    \centering
    \caption{EPC score across different activation functions and data modalities. Bold values indicate the top-performing explainer for each configuration. Missing entries ($-$) represent configurations that were either computationally unfeasible (LIME variants), or where the underlying model failed to converge during training (\textit{ImageNet} with Sigmoid activation). The first row for each model represents the accuracy in the evaluation dataset (see Section \ref{subsec:datasets}).}
    \begin{tabular}{crccc}
         \textbf{Model} & \textbf{Explainer} & \textbf{Bank Loan Default} & \textbf{MNIST} & \textbf{Imagenette} \\
         \hline
         \hline
         \multirow{7}{*}{SiLU} & \textbf{Acc:} & 0.91 & 0.97 & 0.95 \\
         \cline{2-5}
         & Gradient $\times$ Input & 0.49 & 0.90 & 0.86 \\
         & LRP & \textbf{0.50} & 0.92 & 0.88 \\
         & LIG & 0.49 & 0.90 & 0.86 \\
         & IG & \textbf{0.50} & \textbf{0.94} & \textbf{0.92} \\
         & LIME & -0.21 & -0.54 & $-$ \\
         & LIME $\times$ Input & 0.49 & 0.33 & $-$ \\
         \hline 
         \\
         \multirow{7}{*}{tanh} & \textbf{Acc:} & 0.91 & 0.53 & 0.96 \\
         \cline{2-5}
         & Gradient $\times$ Input & 0.59 & 1.52 & 0.66 \\
         & LRP & \textbf{0.60} & 1.74 & \textbf{0.91} \\
         & LIG & 0.59 & 1.52 & 0.66 \\
         & IG & \textbf{0.60} & \textbf{1.77} & 0.83 \\
         & LIME & -0.16 & -0.88 & $-$ \\
         & LIME $\times$ Input & 0.58 & 0.79 & $-$ \\
         \hline
         \\
         \multirow{7}{*}{Sigmoid} & \textbf{Acc:} & 0.89 & 0.86 & 0.16 \\
         \cline{2-5}
         & Gradient $\times$ Input & 0.65 & 0.96 & $-$ \\
         & LRP & 0.08 & 0.40 & $-$ \\
         & LIG & 0.65 & 0.96 & $-$ \\
         & IG & \textbf{0.66} & \textbf{1.08} & $-$ \\
         & LIME & -0.20 & -0.55 & $-$ \\
         & LIME $\times$ Input & 0.65 & 0.48 & $-$ \\
         \hline
    \end{tabular}
    \label{tab:area_vs_problem}
\end{table}

As observed under the smooth, non-monotonic surface of the SiLU activation ($f(x) = x \cdot \sigma(x)$), our results reveal slight variations. Within this regime, LRP slightly outperforms Gradient $\times$ Input in higher dimensions (yielding a EPC score of $0.88$ against $0.86$ in Imagenette). However, IG performs the best across all datasets.

A distinct behavioral shift occurs under the hyperbolic tangent ($\tanh$) activation. Here, LRP and IG share the top ranking (IG ($1.77$) slightly outperforms LRP ($1.74$) on MNIST, whereas LRP ($0.91$) performs better than IG ($0.83$) on Imagenette). However, we observe an additional phenomenon: with the MNIST dataset, the EPC Score exceeds $1.0$ in most cases. This occurs when the model accuracy is relatively low ($0.53$), so nullifying non-relevant attributes improves performance over the baseline, as discussed before, while nullifying relevant attributes drops performance close to zero. As a result, the EPC curve remains above the expected \textit{optimal EPC} ($f(k) = k / 100$), thereby exceeding the expected upper bound and resulting in an EPC score greater than $1$. 

The most critical degradation emerges when evaluating the Sigmoid model. First, the custom CNN failed to learn the Imagenette dataset ($0.16$ evaluation accuracy), leaving that section blank. Second, on the datasets where training converged (Bank Loan Default and MNIST), LRP completely breaks down, with its EPC scores falling to $0.08$ and $0.40$, respectively. This failure empirically validates the theoretical limitation formalized in Section~\ref{subsec:model-specific-methods}: LRP-0 substitutes the activation derivative with the ratio $f(x)/x$. For a Sigmoid layer, as pre-activation inputs approach zero ($x \to 0$), the numerator remains bounded ($\sigma(0) = 0.5$), forcing the pseudo-gradient fraction $\sigma(x)/x$ to explode and causing an uncontrollable propagation of relevance.

In conclusion, these empirical evaluations validate the EPC score as a reliable metric for quantifying explanation quality. It successfully captures the behavioral decay of different explainers, showing that model-agnostic methods underperform compared to gradient-based techniques as problem complexity scales. In addition, Integrated Gradients emerges as one of the most robust explainers according to the EPC. In the following sections, we deepen an analysis of the RNN for sentiment analysis and an exhaustive evaluation of a high-dimensional, state-of-the-art Convolutional Neural Network (MobileNet) under varying baseline configurations, with the aim of determining whether the explainers that perform best according to the EPC score are also preferred under more human-centered evaluation metrics. 

\subsection{Semantic Fidelity under Sequence Modeling}
\label{subsec:results-text}

\begin{figure}
    \centering
    \includegraphics[width=\linewidth]{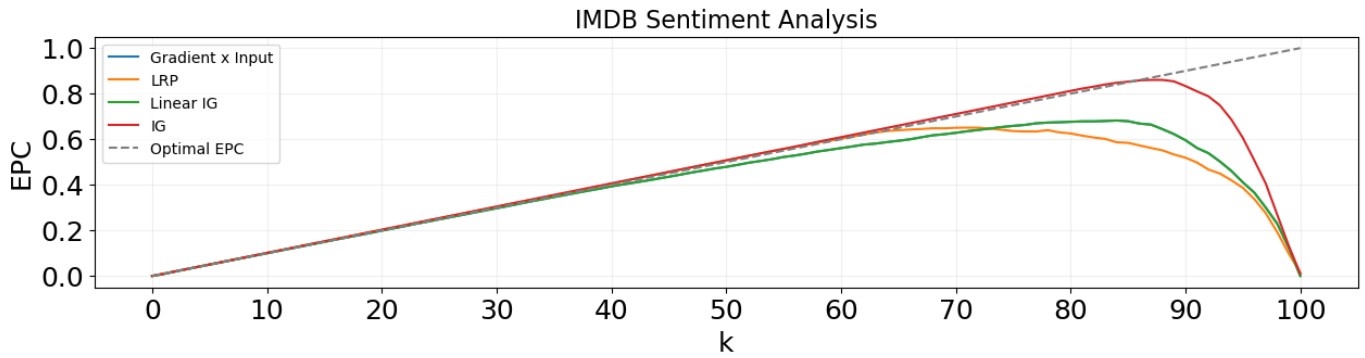}
    \caption{EPC curves for model-specific explainers for IMDB dataset. Note that Gradient $\times$ Input and LIG are equivalent and their curves are overlapping.}
    \label{fig:epc_imdb}
\end{figure}

In the previous section, we evaluated the proposed EPC score as a measure of the quality of explanations. We now explore whether this evaluation is also reflected in human-centered explainability. Specifically, as described in Section \ref{subsec:datasets}, we evaluate the EPC score on the IMDB sentiment analysis dataset, and then we analyze whether explainers with higher EPC scores also exhibit stronger agreement with human semantic knowledge. As described in Section \ref{subsec:models}, we use a standard recurrent architecture with an LSTM. To establish the global explanation fidelity within this problem, we first compute the EPC score. Table~\ref{tab:distance_area_text} and Figure \ref{fig:epc_imdb} show the resulting EPC scores and the EPC curves, respectively, for all model-specific explainers.

\begin{table}[!bht]
    \centering
    \caption{EPC score for sequential token attributions on the IMDB sentiment analysis dataset under an LSTM architecture. Note that Gradient $\times$ Input and LIG are equivalent.}
    \begin{tabular}{r|c|c|c|c}
         \textbf{Explainer:} & Gradient $\times$ Input & LRP & LIG & IG \\
         \hline
         \textbf{EPC score:} & 0.80 & 0.79 & 0.80 & \textbf{0.92} \\
         \hline
    \end{tabular}
    \label{tab:distance_area_text}
\end{table}

As illustrated in the figure and the table, Integrated Gradients (IG) outperforms all the other explainers, yielding an EPC score of $0.92$. In addition, Gradient $\times$ Input and LIG share the same EPC score ($0.80$) due to their equivalence. LRP also achieves almost the same score ($0.79$), but remains in last place. 

This performance gap between explainers raises a qualitative question: do these numerical improvements reflect a superior capacity to extract true linguistic semantics, or are the different methods capturing other structural attributions? To inspect this, we analyze the average attribution assigned to individual words across the validation corpus. Since relevances may have different numerical scales across input texts, directly comparing their absolute values is not meaningful. Instead, our analysis focuses on the relative ranking of tokens within each text. Hence, for each input text, tokens are ranked according to their relevance scores. We then identify the $\gamma$ most relevant tokens (top-$\gamma$) and the $\gamma$ least relevant tokens (bottom-$\gamma$). Each occurrence of a word within the top-$\gamma$ set contributes $+1$ to its score, whereas each occurrence within the bottom-$\gamma$ set contributes $-1$. The contributions are accumulated across the entire evaluation corpus; thus, the accumulated relevance score for a word $w$ is defined as:

\begin{equation}
    R_\gamma(w) = N_{top}(w, \gamma) - N_{bottom}(w, \gamma),
\end{equation}

\noindent where $N_{top}(w, \gamma)$ and $N_{bottom}(w, \gamma)$ denote the number of occurrences of $w$ among the top-$\gamma$ and bottom-$\gamma$ tokens across the evaluation corpus, respectively. To account for differences in word frequency, this accumulated score is normalized by the total number of occurrences of each word in the corpus:

\begin{equation}
    S_\gamma(w) = \frac{R_\gamma(w)}{\text{freq}(w)},
\end{equation}

\noindent where $\text{freq}(w)$ is the word frequency in the corpus. Thus, $S_\gamma(w) \in [-1, 1]$, where positive values refer to words that are consistently selected among the most relevant tokens, while negative values correspond to words that are selected among the least relevant ones.

\begin{figure}
    \centering
    \includegraphics[width=\linewidth]{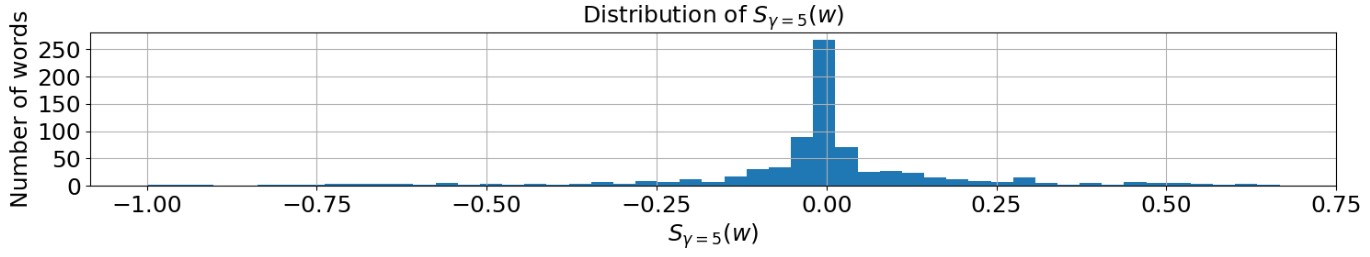}
    \caption{Distribution of $S_{\gamma=5}(w)$ obtained with Integrated Gradients after filtering out words appearing fewer than 20 times in the training corpus. The long-tail distribution indicates that only a small subset of words is consistently selected as the most relevant in both cases. The distributions of the rest of explainers are quite similar, thus not shown.}
    \label{fig:imdb_relevance_distribution}
\end{figure}

However, a raw aggregation uncovers a possible model bias toward low-frequency words. If an unconstrained model encounters an extremely rare token that happens to appear only once or twice in a highly polarized review (e.g., a specific character name like ``Gollum'' inside an overwhelmingly positive review of a fantasy film), the network may overfit and establish an artificial shortcut, associating that rare word with maximum positive sentiment. Thus, to bypass these training artifacts and evaluate whether the explainers capture robust linguistic features, we apply a frequency threshold, filtering out any words with fewer than 20 occurrences. Figure \ref{fig:imdb_relevance_distribution} shows the resulting distribution of $S_{\gamma=5}(w)$.

The distribution shows a pronounced long-tail behavior: the majority of words receive values close to zero and only a small number of tokens are consistently selected among the most (or least) relevant across many samples accumulating large positive (or negative) scores. This indicates that the explanations are highly concentrated on a small vocabulary of sentiment-bearing words, whereas most words only sporadically appear among the relevance rankings.

After identifying the words that are most consistently assigned high and low relevance, we evaluate whether these explanations agree with human-defined notions of sentiment polarity. To this end, we compare the extracted word rankings against the AFINN sentiment lexicon \cite{AFINN_Nielsen2011}, a manually annotated lexical resource in which English words are assigned integer sentiment scores ranging from $-5$ (strongly negative) to $+5$ (strongly positive). AFINN provides word-level polarity annotations independent of contextual information, making it particularly suitable for evaluating whether attribution methods consistently identify words that humans perceive as carrying positive or negative sentiment.

Table~\ref{tab:high_freq_relevances} presents the 10 words with the highest and lowest score $S_{\gamma=5}(w)$ for the Gradient $\times$ Input and IG  explainers, together with their corresponding AFINN sentiment scores. We expect words that are consistently selected among the top-5 to exhibit positive sentiment scores, whereas words frequently appearing in the bottom-5 should receive negative scores.

\begin{table}[!bht]
    \centering
    \caption{Top $\gamma=5$ most and least relevant words under a frequency filter ($\text{freq} > 20$). We show in the table the normalized score $S_{\gamma=5}(w)$ and its AFINN sentiment score (A). Bold text highlights words carrying strong explicit semantic alignment with AFINN.}
    \begin{tabular}{ccc|ccc}
        \multicolumn{6}{c}{\textbf{Most relevant words}}  \\
        \multicolumn{3}{c}{Gradient $\times$ Input} & \multicolumn{3}{c}{Integrated Gradients} \\
        Word & $S_{\gamma=5}(w)$ & A & Word & $S_{\gamma=5}(w)$ & A \\
        \hline
        7 & 0.52 & 0
        & 7 & 1.0 & 0 \\
        hot & 0.45 & 0
        & \textbf{ enjoyable } & 0.93 & 2 \\
        \textbf{ fun } & 0.45 & 4
        & \textbf{ excellent } & 0.89 & 3 \\
        \textbf{ perfect } & 0.42 & 3
        & unique & 0.83 & 0 \\
        highly & 0.38 & 0
        & \textbf{ loved } & 0.78 & 3 \\
        \textbf{ enjoyable } & 0.37 & 2
        & today & 0.78 & 0 \\
        \textbf{ entertaining } & 0.37 & 2
        & \textbf{ glad } & 0.78 & 3 \\
        present & 0.36 & 0
        & \textbf{ wonderful } & 0.78 & 4 \\
        number & 0.36 & 0
        & \textbf{ perfect } & 0.77 & 3 \\
        today & 0.35 & 0
        & highly & 0.76 & 0 \\
        \hline 
        & \textbf{Sum:} & 11 & & \textbf{Sum:} & 18  \\
        \hline
    \end{tabular}

    \vspace{0.5cm}
    
    \begin{tabular}{ccc|ccc}
        \multicolumn{6}{c}{\textbf{Least relevant words}}  \\
        \multicolumn{3}{c}{Gradient $\times$ Input} & \multicolumn{3}{c}{Integrated Gradients} \\
        Word & $S_{\gamma=5}(w)$ & A & Word & $S_{\gamma=5}(w)$ & A \\
        \hline
        \textbf{ avoid } & -0.65 & -1
        & \textbf{ worst } & -0.9 & -3 \\
        \textbf{ waste } & -0.57 & -1
        & predictable & -0.85 & 0 \\
        save & -0.57 & 2
        & \textbf{ waste } & -0.85 & -1 \\
        \textbf{ dull } & -0.56 & -2
        & \textbf{ boring } & -0.8 & -3 \\
        \textbf{ shame } & -0.5 & -2
        & \textbf{ avoid } & -0.8 & -1 \\
        ms & -0.5 & 0
        & \textbf{ disappointing } & -0.8 & -2 \\
        \textbf{ disappointing } & -0.5 & -2
        & \textbf{ awful } & -0.79 & -3 \\
        basically & -0.42 & 0
        & \textbf{ poorly } & -0.78 & -2 \\
        \textbf{ boring } & -0.41 & -3
        & \textbf{ pointless } & -0.75 & -2 \\
        wonder & -0.41 & 0
        & save & -0.74 & 2 \\
        \hline 
        & \textbf{Sum:} & -9 & & \textbf{Sum:} & -15 \\
        \hline
    \end{tabular}
    \label{tab:high_freq_relevances}
\end{table}

The results shown in the table provide qualitative evidence supporting the proposed EPC score. Although both methods identify several sentiment-bearing words, their rankings differ in semantic coherence. For the IG explainer, which obtains the highest EPC score, the top- and bottom- ranked terms accumulate positive and negative polarities of $+18$ and $-15$, respectively, clearly stronger than those ($+11$ and $-9$) obtained for the terms selected by the Gradient $\times$ Input explainer (which in turn was assigned a much smaller EPC score). This indicates that IG concentrates its attribution on more explicitly sentiment-bearing words, and supports our claim that the EPC selects explainers that are more aligned with human-centered evidence. Similar results are obtained for the rest of explainers, although they are not shown due to a lack of space.

\begin{figure}
    \centering
    \includegraphics[width=\linewidth]{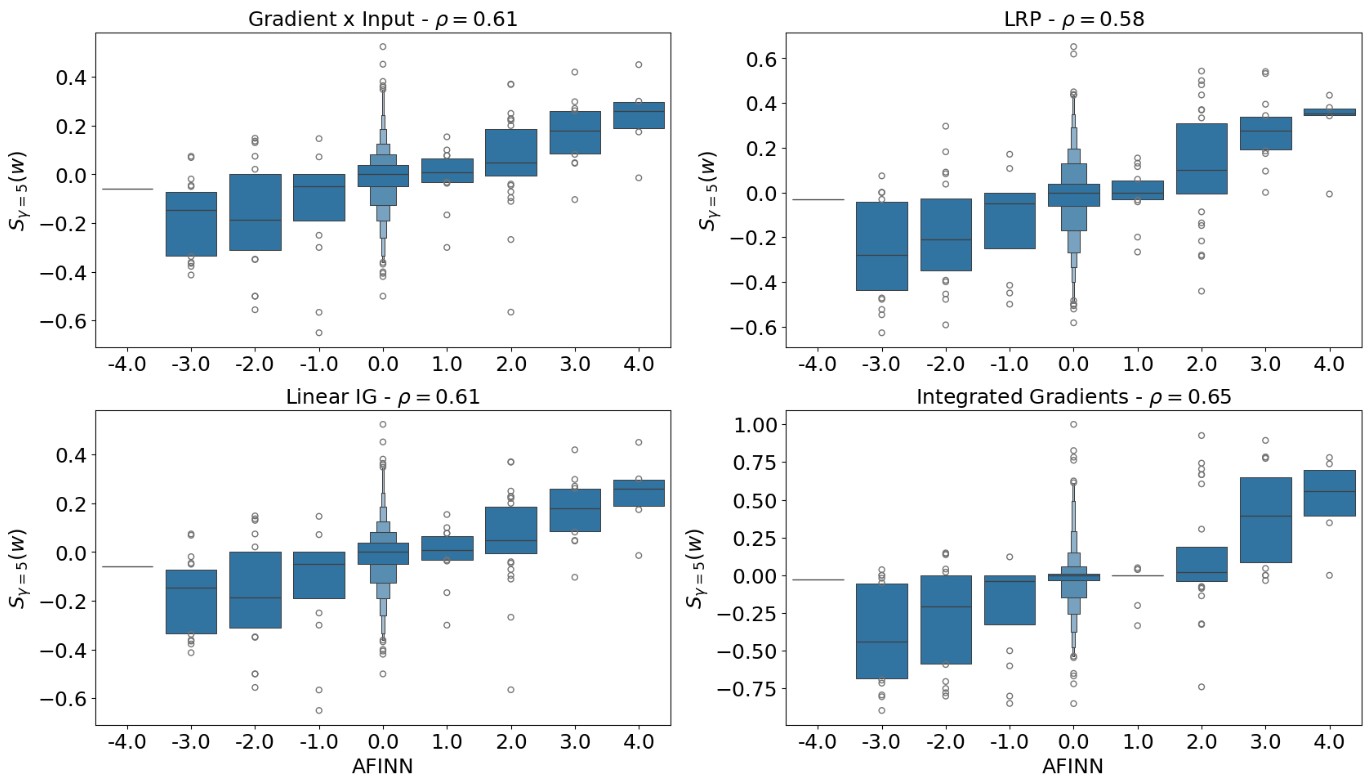}
    \caption{Boxen plots comparing the score $S_{\gamma=5}(w)$ against the corresponding AFINN sentiment score for different explainers. The panels show, from left to right, Gradient $\times$ Input, LRP, Linear IG, and IG. Each subplot reports the Pearson correlation coefficient $\rho$ after excluding words with $|S_{\gamma=5}(w)| < 0.25$.}
    \label{fig:imdb_corr}
\end{figure}

While the qualitative inspection of representative words provides intuitive evidence, we seek a quantitative measure of semantic agreement. Thus, we compare the score $S_{\gamma=5}(w)$ with its corresponding AFINN sentiment score by computing the Pearson correlation over the vocabulary. Since many words never appear among either the top-$\gamma$ or bottom-$\gamma$ relevant tokens, they receive a score of zero. Including these words introduces a large number of semantically uninformative samples. Therefore, we report the correlation after excluding words whose occurrence is lower than 25\% ($|S_{\gamma=5}(w)| < 0.25$). Figure \ref{fig:imdb_corr} presents the resulting boxen plots together with the corresponding Pearson correlation coefficients ($\rho$) for each explainer, illustrating the degree of alignment between our score $S_{\gamma=5}(w)$ and the sentiment polarity assigned by AFINN.

The results shown in the figure confirm the observations discussed above. Integrated Gradients achieves the highest agreement with the AFINN lexicon ($\rho=0.65$), followed by Linear IG and Gradient $\times$ Input ($\rho=0.61$), and LRP ($\rho=0.58$). This ordering exactly matches the ranking induced by the EPC score. Thus, explainers with higher EPC scores also exhibit stronger agreement with human lexical judgments. Overall, these findings suggest that the proposed EPC score captures a notion of explanation fidelity aligned with independently constructed human lexical resources. 

\subsection{Visual Fidelity under Image Classification}
\label{subsec:results-vision}

To further assess the proposed EPC score beyond sequence modeling, we next consider the image classification domain using the pre-trained \textit{MobileNet} architecture \cite{mobilenet_2017}. In this case, evaluating visual explanations requires defining how relevant pixels are removed from the input image. As discussed in Section~\ref{subsec:nullification}, the nullification strategy affects the fidelity evaluation, since different strategies may introduce distinct image artifacts. Consequently, we first analyze the influence of the nullification operator on the EPC score, and then explore whether the resulting EPC rankings are also consistent with human visual annotations obtained from \textit{ImageNet} region-of-interest (ROI) dataset. 

\begin{figure}[!bht]
    \centering
    \includegraphics[width=0.49\linewidth]{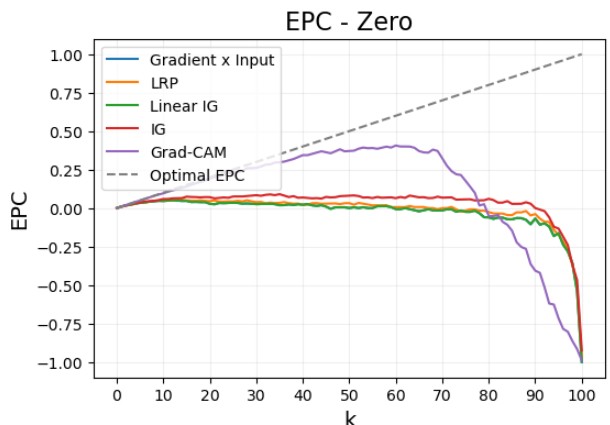}
    \includegraphics[width=0.49\linewidth]{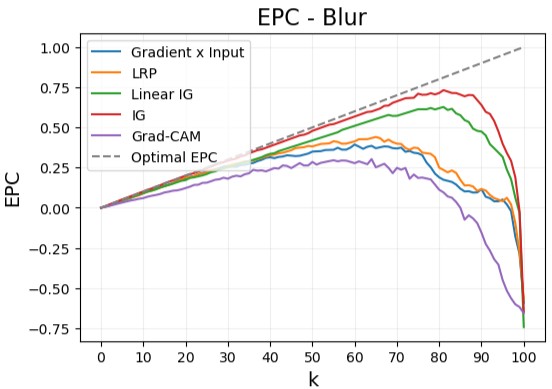}
    \caption{EPC curves comparison on MobileNet using the \textit{ImageNet} subset: constant zero nullification (left) versus Gaussian blur (right).}
    \label{fig:vision-macro-curves}
\end{figure}

To evaluate how the EPC responds to the nullification strategy, we compare zero-masking and blur baselines. In addition to the gradient-based methods considered in the previous sections, we include Grad-CAM, a region-based explainer specifically designed for convolutional neural networks (see Section \ref{subsec:model-specific-methods}). Specifically, the blur operator applies an isotropic Gaussian filter configured with a kernel size of $(15, 15)$, clipping the outputs back to the valid $[-1, 1]$ using the OpenCV library \cite{opencv_library}. Figure~\ref{fig:vision-macro-curves} contrasts the EPC curves of the explainers under the zero-masking regime (left) with the smooth Gaussian blur baseline (right), while Table~\ref{tab:vision-epc-score} summarizes the resulting EPC scores.

\begin{table}[!bht]
    \centering
    \caption{EPC scores obtained on the \textit{ImageNet} dataset using two different nullification strategies. Constant zero masking introduces artificial image discontinuities that favor coarse-grained explanation methods (Grad-CAM), whereas Gaussian blur preserves natural image statistics and enables a fair comparison of pixel-level explainers.}
    \begin{tabular}{cccccc}
         & \multicolumn{5}{c}{$EPC Score$} \\
        \cline{2-6}
        & Gradient $\times$ Input & LRP & LIG & IG & Grad-CAM \\
        \hline
        Zero-nullification & -0.05 & -0.02 & -0.05 & 0.06 & \textbf{0.21} \\
        Blur-nullification & 0.41 & 0.47 & 0.68 & \textbf{0.80} & 0.20 \\
        \hline
    \end{tabular}
    \label{tab:vision-epc-score}
\end{table}

These results reveal that the nullification operator has an interesting impact on the explanation fidelity. Under conventional zero masking, Grad-CAM obtains the highest EPC score of $0.21$, while fine-grained explainers perform poorly despite their pixel-level resolution. This behavior does not necessarily indicate superior explanations, but rather reflects an evaluation bias introduced by the perturbation itself. 

\begin{figure}[!bht]
    \centering
    \includegraphics[width=\linewidth]{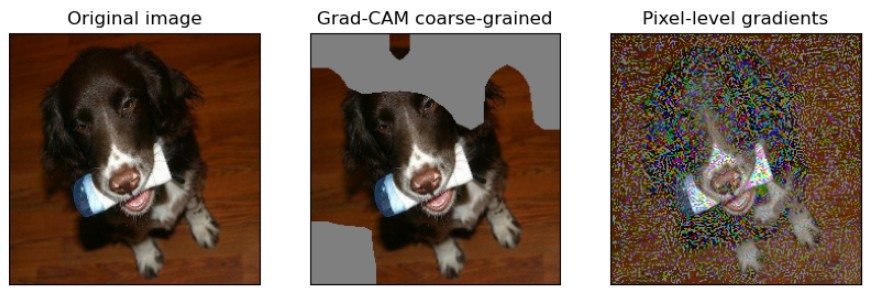}
    \caption{Visual illustration of structural degradation induced by zero-masking on a \textit{MobileNet} instance after nullifying the $k=30\%$ of least relevant features. Left: Original image. Middle (Grad-CAM): coarse-grained regional patching that masks contiguous spatial blocks, inadvertently protecting the architecture from severe out-of-distribution artifacts. Right (pixel-level gradients): pixel-guided deletion resulting in a highly fractured, fine-grained noise matrix that triggers immediate adversarial confidence drops.}
    \label{fig:mobilenet-grad-vs-patch}
\end{figure}

As illustrated in Figure~\ref{fig:mobilenet-grad-vs-patch}, Grad-CAM generates coarse, spatially contiguous attribution regions (middle figure). Consequently, removing its most relevant features replaces relatively compact image patches with a uniform baseline, introducing few artificial edges. In contrast, pixel-level explainers identify sparse and highly localized relevance patterns (Figure~\ref{fig:mobilenet-grad-vs-patch}-right). Zero masking these sparse pixels produces a large number of isolated discontinuities that substantially alter the image distributions, causing the classifier confidence to decrease even when non-relevant pixels are removed. The resulting EPC scores underestimate the fidelity of fine-grained explanation methods.

However, when the nullification operator is switched to the Gaussian blur strategy, the structural bias is dismantled, aligning with the benchmark warnings highlighted by Gomez et al. \cite{gomez2022}. In this scenario, the evaluation better reflects the informational content of the explanations by largely removing the structural artifacts introduced by zero masking. Under this setting, explainers producing fine-grained relevance rankings achieve substantially higher EPC scores than coarse regional attribution methods: IG achieves a commanding peak performance ($0.80$), followed by Linear IG ($0.68$), while Gradient $\times$ Input and LRP hover around $0.41$ and $0.47$, respectively. In the last position, Grad-CAM gets an EPC score of $0.20$. This reversal suggests that the superior performance of Grad-CAM under zero masking is attributable to the perturbation operator rather than the quality of its explanations. Once this structural bias is removed, the EPC consistently favors explainers that provide a more discriminative relevance ranking. This behavior is closely related to the ability of pixel-level methods to distinguish necessary evidence from dispensable evidence, rather than simply maximizing the spatial extent of the highlighted object.

\begin{figure}
    \centering
    \includegraphics[width=\linewidth]{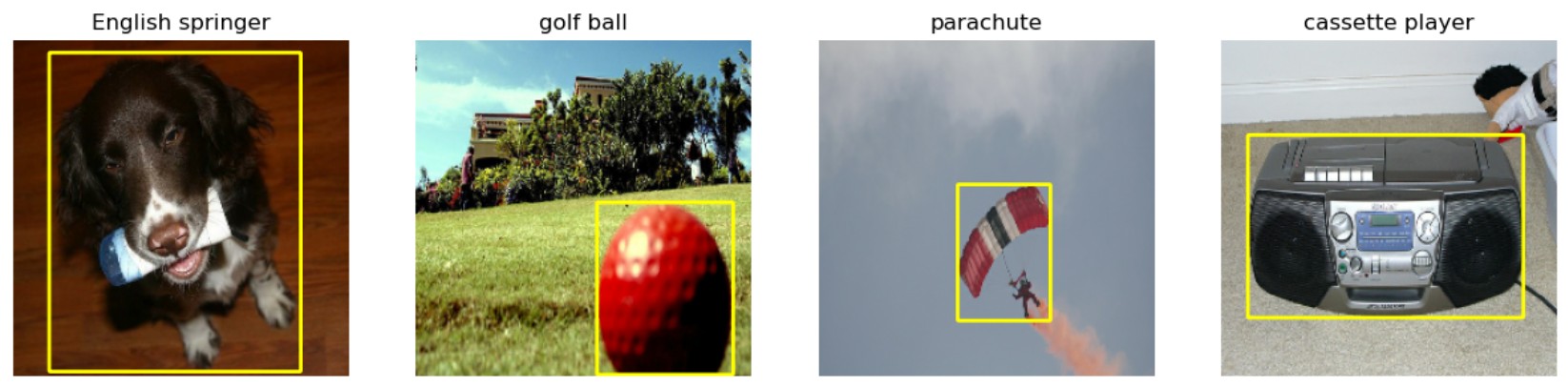}
    \caption{Examples of images from the \textit{ImageNet} Object Localization dataset together with their manually annotated Regions of Interest (ROIs). The yellow rectangles denote the human-provided bounding boxes identifying the primary object used as spatial ground truth in our evaluation.}
    \label{fig:examples_roi}
\end{figure}

After analyzing the behavior of the EPC score under different nullification strategies, we now evaluate whether the observed differences are also reflected in an independent measure of human understandability. To this end, we use the Region of Interest (ROI) annotations provided by the \textit{ImageNet} localization set \cite{object_localization_ILSVRC15}, publicly available through its Kaggle repository \cite{imagenet-object-localization-challenge}. These annotations manually delimit the spatial location of the main object present in each image. Figure \ref{fig:examples_roi} illustrates several representative examples together with their corresponding annotated ROIs. 

Our hypothesis is analogous to that explored in the previous text classification experiment. If the proposed EPC score captures explanation fidelity, explainers obtaining higher EPC scores should also focus their relevance on image regions that humans identify as containing the object of interest. Consequently, we explore the relationship between the EPC score and the spatial agreement with the annotated ROIs.

To quantify this agreement, we cast the evaluation as a binary segmentation problem: the annotated ROI defines the positive class, corresponding to the object of interest, while all pixels outside the annotation are treated as background and therefore constitute the negative class. For a given value of $\gamma$, we retain only the pixel subsets $Z_\gamma^{top}$ with the top-$\gamma\%$ highest relevance values, and $Z_\gamma^{bottom}$ with the bottom-$\gamma\%$ lowest relevance values to obtain a binary prediction mask. This mask is then compared pixel-wise against the annotated ROI, yielding the corresponding confusion matrix from which Precision is computed: 

\begin{equation}
    Prec^{top} = \frac{|Z_\gamma^{top} \cap ROI|}{|Z_\gamma^{top}|}
\end{equation}

\begin{equation}
    Prec^{bottom} = \frac{|Z_\gamma^{bottom} \cap \overline{ROI}|}{|Z_\gamma^{bottom}|}
\end{equation}

For the top-$\gamma$ evaluation, the annotated ROI defines the positive class, as the highest-ranked relevance values are expected to identify the object of interest. For the bottom-$\gamma$ evaluation, the positive class is instead defined by the background (the complement of the annotated ROI) since the lowest-ranked relevance values are expected to identify regions that are not required for the prediction.

The parameter $\gamma$ determines the fraction of the relevance ranking under evaluation. We focus our analysis by fixing $\gamma=1\%$ ($1.506$ features). This choice isolates the most critical relevance decisions made by each explainer: $Top@1$ measures whether the highest-ranked pixels ($Z_1^{top}$) succeed in capturing the most necessary evidence inside the human-annotated ROI, whereas $Bottom@99$ assesses whether the lowest-ranked pixels ($Z_1^{bottom}$) are correctly assigned to dispensable background regions. If the proposed EPC score reflects explanation fidelity, explainers achieving higher EPC values (IG and LIG with blur-nullification) should also exhibit superior precision at this fine-grained threshold. Table \ref{tab:bac_top_bottom_extremos} summarizes the EPC scores alongside the precision obtained for all evaluated explainers.

\begin{table}[!bht]
    \centering
    \caption{EPC score and Precision measured at two representative operating points of the relevance ranking. \textit{Top@1} and \textit{Bottom@99} evaluate the two ranking, where explainers must identify the most necessary and most dispensable evidence, respectively.}
    \begin{tabular}{cccc}
         & & \multicolumn{2}{c}{\textbf{Precision}} \\
         \textbf{Explainer} & \textbf{EPC} & $Top@1$ & $Bottom@99$ \\
         \hline
         \hline
         Gradient $\times$ Input & 0.42 & 0.70 & 0.55 \\
         LRP & 0.47 & 0.71 & 0.53 \\
         LIG & 0.68 & \textbf{0.79} & \textbf{0.56} \\
         IG & 0.8 & 0.78 & \textbf{0.56} \\
         Grad-CAM & 0.2 & 0.65 & 0.55 \\
         \hline
    \end{tabular}
    \label{tab:bac_top_bottom_extremos}
\end{table}

As shown in the table, the pixel-level explainers obtain comparatively higher scores with the precision $Top@1$. Both IG variants ($0.78-0.79$) outperform simpler fine-grained explainers ($0.70-0.71$) and Grad-CAM ($0.65$). Here, the differences are substantially larger because the evaluation focuses on a much smaller positive region, making precision more discriminative. In the other scenario, $Bottom@99$, although the absolute differences are relatively small because the background occupies most of the image, the ordering of explainers coincides again, being both IG variants ($0.56$) the most precise explainers.

This behavior can be explained by the different spatial characteristics of the relevance maps produced by the evaluated explainers. Grad-CAM typically generates smooth and spatially coherent activation maps that cover large portions of the object. Consequently, Grad-CAM reflects object localization rather than the precise ordering of individual pixels. In contrast, pixel-level explainers produce much more selective relevance maps. Rather than assigning high relevance across the entire object, they concentrate attribution on a relatively small set of highly discriminative pixels in the object regions. 

\begin{figure}[!bht]
    \centering
    \includegraphics[width=0.48\linewidth]{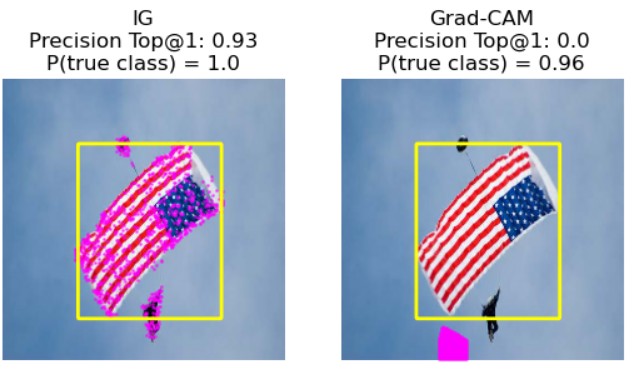}
    \includegraphics[width=0.48\linewidth]{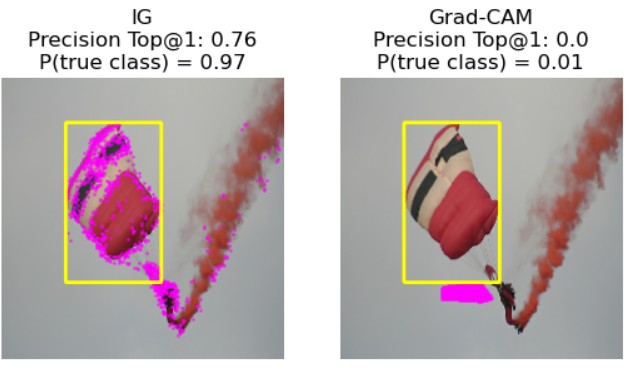} \\
    \includegraphics[width=0.48\linewidth]{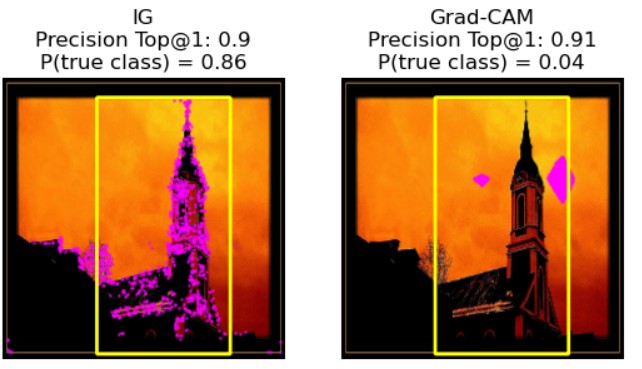}
    \includegraphics[width=0.48\linewidth]{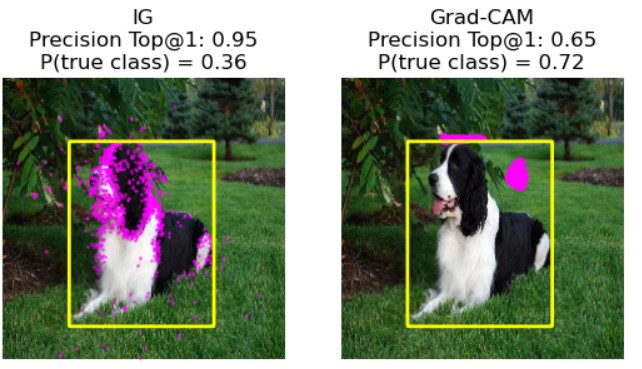}
    \caption{Illustrative comparison between Integrated Gradients (IG) and Grad-CAM at $Top@1$ ($\gamma=1\%$) across four representative cases. High spatial agreement with the human ROI does not always translate to high predictive confidence when retaining selected pixels. This highlights the complementary nature of human-annotated ROI localization versus fidelity-based evaluation measured by the EPC.}
    \label{fig:illustration-of-coarse-vs-fine-ROI}
\end{figure}

This distinction is illustrated in Figure~\ref{fig:illustration-of-coarse-vs-fine-ROI}, where we compare several predictions explained by IG and Grad-CAM at $\gamma = 1$ with different examples. To evaluate whether the highlighted pixels actually contain the predictive evidence, we show both the spatial precision with the ROI (Precision $Top@1$) and the model's confidence when retaining only those top $1\%$ pixels.

As shown across these four scenarios, high spatial agreement with human annotations does not guarantee that an explainer has isolated the true evidence used by the network:

\begin{itemize}
    \item \textbf{Context Dependence (Top-Left):} In the american parachutist example, IG achieves $Prec@1 = 0.93$ by highlighting pixels strictly across the parachute canopy, yielding an absolute model confidence ($1.00$). Conversely, Grad-CAM scores $Prec@1 = 0.0$ because its activation focuses entirely on the blue sky. Interestingly, retaining only those blue sky pixels still yields $0.96$ of confidence in the target class, exposing a dataset bias where the background context alone allows the classifier to infer the target class.
    
    \item \textbf{Contextual Failure (Top-Right):} In a second parachutist image with a cloudy grey sky, IG retains high precision ($Prec@1 = 0.76$) and high confidence ($0.97$) by correctly selecting pixels on both the canopy and the jumper. In contrast, Grad-CAM again targets the sky background ($Prec@1 = 0.0$), but because the sky lacks the blue contextual cue, the classifier confidence collapses ($0.01$). This proves that Grad-CAM's selection was spurious rather than necessary for the decision.
    
    \item \textbf{Failure within ROI (Bottom-Left):} In the building instance at sunset, both methods achieve high spatial overlap with the human ROI ($Prec@1 = 0.90$ for IG and $0.91$ for Grad-CAM). However, while IG concentrates its top pixels on the building structure itself ($0.86$), Grad-CAM highlights the orange sky enclosed inside the bounding box, leading to a complete prediction failure ($0.04$). High ROI precision can thus mask incorrect explanations.
    
    \item \textbf{Spurious Grass Cues (Bottom-Right):} In the dog instance, IG selectively targets facial features ($Prec@1 = 0.95$, and $P = 0.3555$). Grad-CAM achieves lower spatial precision ($Prec@1 = 0.65$) by focusing largely on surrounding grass patches rather than the animal itself, yet retains higher confidence ($0.72$), indicating that the model relies heavily on environmental textures.
\end{itemize}

These examples illustrate that spatial agreement with the ROI alone is insufficient to determine whether the selected pixels correspond to the evidence actually required by the model. Pixel-level explainers, such as IG, successfully distinguish the object regions that are necessary for the prediction from those that are merely part of the object but are largely dispensable to the model's decision. ROI annotations do not distinguish between necessary and dispensable evidence within the object itself. 

This observation is closely related to the objective of the EPC score. Rather than rewarding explainers that maximize spatial coverage of the annotated object, the EPC evaluates whether the relevance ranking successfully separates necessary evidence from dispensable evidence according to the model. At \textit{Top@1}, the EPC asks whether the highest-ranked pixels correspond to the most necessary evidence for the prediction, whereas at \textit{Bottom@99} it asks whether the lowest-ranked pixels are assigned to genuinely dispensable regions. This behavior is consistent with the results observed in Table~\ref{tab:bac_top_bottom_extremos}. 

In other words, object localization and relevance ranking are not competing objectives but complementary ones. The former evaluates where the evidence lies, whereas the latter evaluates which parts of that evidence are actually required by the model. This agreement is consistent with the objective of EPC. Since the metric evaluates how well the necessary evidence is presented, its behavior depends on the quality of the relevance ordering. The EPC requires a fine-grained ordering of individual pixels and therefore favors selective pixel-level explainers, such as IG variants, over coarse explainers (Grad-CAM).

Overall, these results provide strong evidence that the EPC captures a meaningful notion of visual explanation fidelity. Although the EPC is not designed to measure object localization, it shows a clear correspondence with ROI-based evaluation, where the correct ordering of individual pixels becomes the determining factor. This agreement indicates that the EPC successfully measures whether an explainer identifies the evidence that is necessary for the model prediction while separating it from evidence that is largely dispensable. From this perspective, the experiments provide empirical evidence that human annotations validate explanations only up to the level of object localization. EPC extends this validation by assessing whether the model actually relies on the annotated evidence for its prediction. 

Among the evaluated explainers, the IG family consistently achieves the highest EPC scores while maintaining agreement with the annotated ROIs. Pixel-level attribution methods produce a more informative relevance ranking than coarse localization methods.

\section{Conclusions and Future Work}
\label{sec:conclusions}

In this work, we proposed the EPC score, a scalar extension of the Explainability-Performance Coefficient (EPC) introduced by Oliva and Lago-Fernández \cite{oliva25}. While the original EPC evaluates explanation quality across continuous curves, our proposed score compresses this behavior into a single numerical metric per explainer. This allows for direct and quantitative comparisons among explainability methods across diverse neural architectures (MLPs, CNNs, LSTMs) and data modalities (tabular, image, and text). 

Across an experimental pipeline from simple benchmarks to complex tasks such as sentiment analysis on IMDB and image classification on \textit{ImageNet}, we analyzed both model-specific (gradient-based) and model-agnostic explainers under varying activation functions and input nullification strategies. Our results yield two main conclusions regarding explanation quality and human alignment. First, the proposed EPC score serves as a reliable metric for evaluating explainer fidelity. Model-agnostic methods (LIME, SHAP) scale poorly in both computational complexity and attribution quality as input dimensionality grows. Conversely, gradient-based approaches, specifically Integrated Gradients (IG), outperform simpler explainers and model-agnostic alternatives across all modalities and activation functions. Furthermore, our image classification experiments demonstrate that evaluating visual fidelity is highly sensitive to the chosen nullification strategy. Blurring eliminates structural artifact biases and highlights the fine-grained precision of pixel-level explainers.

Second, the EPC score has strong consistency with independent human-annotated ground truths. In sentiment analysis, higher EPC scores strongly correlate with human lexical judgments from the AFINN sentiment dictionary ($\rho = 0.65$ for IG). In visual domains, higher EPC scores reflect superior precision within human-annotated Regions of Interest (ROI). In addition, our results reveal a conceptual distinction between human visual annotations and computational fidelity. Thus, object localization and explanation fidelity must be understood as complementary rather than equivalent dimensions of explainability. 

Several promising avenues remain open for future research. Although Gaussian blur acts as an effective and computationally light proxy to prevent artificial image discontinuities, future work will investigate generative inpainting techniques, such as Diffusion models, to reconstruct nullified regions, further minimizing out-of-distribution artifacts. Additionally, we intend to extend the EPC score to large-scale architectures, including Large Language Models (LLMs) and Vision-Language Transformers, where multi-head attention mechanisms present unique structural and alignment challenges for attribution fidelity.

\section*{Acknowledgements}

This research was supported by grant PID2023-149669NB-I00 (MCIN/AEI and ERDF - ``A way of making Europe'').

%
%
%
\bibliographystyle{splncs04}
\bibliography{mybib}

\end{document}